\documentclass[journal]{IEEEtran}
\usepackage{latexsym, epsfig, amssymb, graphics, graphicx, amsmath, booktabs, comment, arydshln, dsfont, algpseudocode}

\def\L{\mathcal{L}}
\newcommand{\R}{\mathbb{R}}

\def\e{\varepsilon} 

\begin{document}

\title{Spiking Manifesto}

\author{Eugene Izhikevich, SpikeCore, California}


\maketitle

\begin{abstract}

Practically everything computers do is better, faster, and more power-efficient than the brain. For example, a calculator performs numerical computations more energy-efficiently than any human. Yet modern AI models are a thousand times less efficient than the brain. These models rely on larger and larger artificial neural networks (ANNs) to boost their encoding capacity, requiring GPUs to perform large-scale matrix multiplications. In contrast, the brain’s spiking neural networks (SNNs) exhibit factorially explosive encoding capacity and compute through the polychronization of spikes rather than explicit matrix–vector products, resulting in lower energy requirements. This manifesto proposes a paradigm for framing popular AI models in terms of spiking networks and polychronization, and for interpreting spiking activity as nature's way of implementing look-up tables. This suggests a path toward converting AI models into a novel class of architectures with much smaller size yet combinatorially large encoding capacity, offering the promise of a thousandfold improvement in performance. 
Code is available at https://github.com/izhikevich/SNN
\end{abstract}

\section{The manifesto}
\begin{figure}[t]
    \centering
    \includegraphics[width=0.44\textwidth]{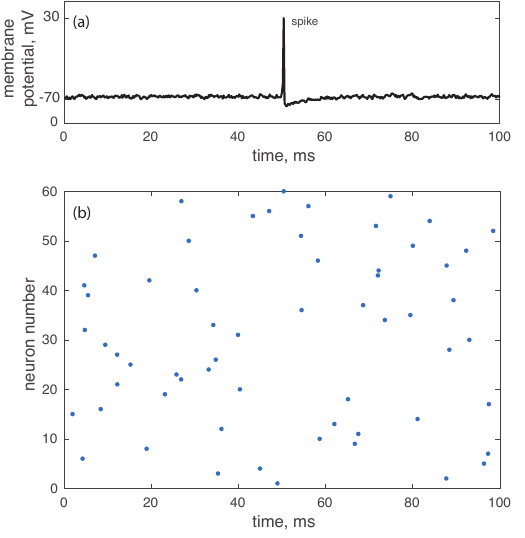}
    \caption{(a) The all-or-none spiking response of a cortical neuron to a random input. (b) 60 neurons, each firing only one spike, have a representation capacity 60!, which is greater than the number of particles in the visible universe. }
    \label{60 neurons}
\end{figure}

Llama 3 has artificial neurons, $10^{11}$ (synaptic) parameters, and consumes tens of watts in inference mode. In contrast, the human brain has spiking neurons, $10^{15}$ synapses, and consumes only 20 watts. Scaled down 10,000-fold to the size of Llama 3, the brain would require only 2 mW -- enough to power it for a month with a single AAA battery! There are two reasons for the power efficiency of the brain: 

{\sc Efficient processing:} Spiking neural networks (SNNs) of the brain have sparse activity. If the input to a spiking neuron is below its threshold, the neuron is silent and no signal is transmitted. Only when the input exceeds the threshold the neuron generates a brief all-or-none event, called {\em spike} in Fig. \ref{60 neurons}a, that propagates to target neurons. This signaling sparsity provides unparalleled efficiency. In contrast, artificial neurons in popular AI architectures generate continuous-valued neuronal activations. At each time step, activations are transmitted to target neurons and multiplied by synaptic weights to calculate new inputs, resulting in a massive demand for GPU matrix multiplications. 

{\sc Efficient encoding:} Neuronal activations in ANNs are often interpreted as probabilities of spiking or as instantaneous firing rates. Both interpretations ignore the timing of spikes and miss the opportunity to have a factorially explosive representation capacity. In Fig. \ref{60 neurons}b, we depict 60 neurons, each firing only 1 spike during a certain time period. From an ANN's point of view, this corresponds to all activations being equal to $1$, that is, to a single feature vector, a single embedding, with activation values $(1, 1, \ldots, 1) \in\R^{60}$. Changing the order of spikes does not change the activation values. However, from an SNN's point of view, this is just one pattern out of $60$ factorial possible patterns of different relative spiking orders. 60! is greater than $10^{80}$ -- the number of particles in the visible universe. Therefore, a single representation value from an ANN's point of view has a practically infinite representation capacity from an SNN's point of view. 

\begin{figure}[t]
    \centering
    \includegraphics[width=0.43\textwidth]{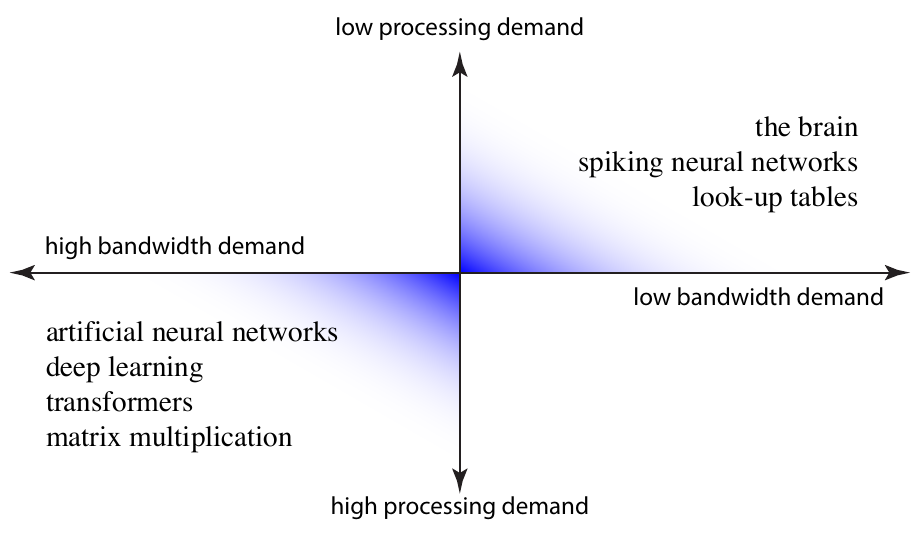}
    \caption{Classification of ANNs and SNNs based on the "memory vs. processing" (space vs. time) demand. SNNs typically have large memory footprint, but only a tiny fraction is used at any given forward pass, resulting in low memory bandwidth requirement. }
    \label{memory-processing}
\end{figure}

High representational capacity reduces memory requirements, and sparse activity reduces computations, placing SNNs in the quadrant opposite to ANNs in the memory–processing diagram in Fig. \ref{memory-processing}. However, current SNN research rarely goes beyond sparsity. This manifesto calls for fully harnessing the combinatorially explosive representational capacity of SNNs to claim and dominate the top-right quadrant.

\begin{figure}[t]
    \centering
    \includegraphics[width=0.46\textwidth]{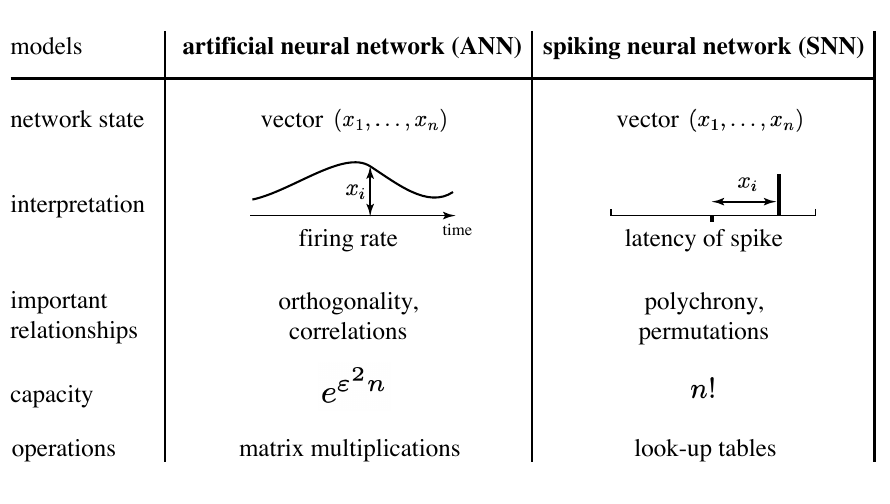}
    \caption{The distinction between ANNs and the SNN described in this manifesto. The linear capacity $e^{\varepsilon^2 n}$ to represent $\varepsilon$-orthogonal vectors \cite{JohnsonLindenstrauss}, is dwarfed by the combinatorial capacity $n!$.}
    \label{summary_of_features}
\end{figure}

One could argue that the interpretations of neuronal activations do not matter because the states of both ANNs and SNNs can be described by vectors encoding features; indeed, the spike-timings in Fig\;\ref{60 neurons}b can be described by a 60-dimensional vector of spike latencies. However, the meaning we assign to these vectors fundamentally changes how we operate on them, summarized in Fig. \ref{summary_of_features}. In ANNs, the interpretation of vector values as continuous activations utilizes relationships like orthogonality and correlation, and relies heavily on operations such as dot products and matrix multiplications. This computational paradigm is what allowed ANNs to win the GPU hardware lottery \cite{Hardware Lottery}, as GPUs are highly optimized for these exact operations. 

\begin{figure}[t]
    \centering
    \includegraphics[width=0.5\textwidth]{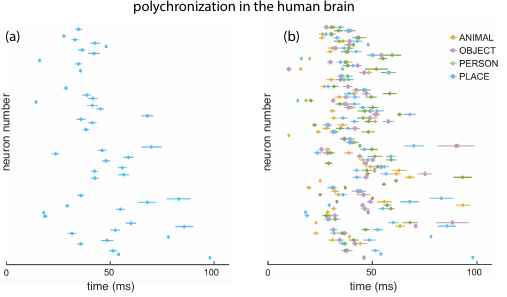}
    \caption{(a) Precise and repeatable sequence of spikes (with error bars) in {\em in vivo} recordings of 64 human anterior temporal lobe neurons while presenting visual images of places. (b) The same neurons polychronize with different repeatable sequences while the subject is viewing other image categories \cite{XieNature}. {\em Syn}-chronous -- one timing; {\em poly}-chronous -- multiple timings \cite{Polychronization}.  }
    \label{polychronous_patterns.pdf}
\end{figure}

\looseness=-1
This manifesto treats spiking patterns as vectors of latencies. This interpretation makes orthogonality, correlations, and matrix multiplications irrelevant. Instead, features are represented by {\em polychronous spiking} patterns, illustrated in Fig. \ref{polychronous_patterns.pdf}, where the precise order of spikes within each pattern is important. Polychronization was first observed in biologically-detailed brain models \cite{CerCor, PNAS} and then in the human brain \cite{Vaz2020,Vaz2023,XieNature}. These patterns represent memories and their number is often greater than the number of synapses in the network \cite{Polychronization}. However, simulating spiking networks requires modeling membrane potentials of individual spiking neurons and direct interactions between such neurons - a monumental task best suited for traditional neuromorphic hardware like TrueNorth \cite{TrueNorth}, LOIHI \cite{Loihi2}, or SpiNNaker \cite{Spinnaker2}. 

\begin{figure}[t]
    \centering
    \includegraphics[width=0.45\textwidth]{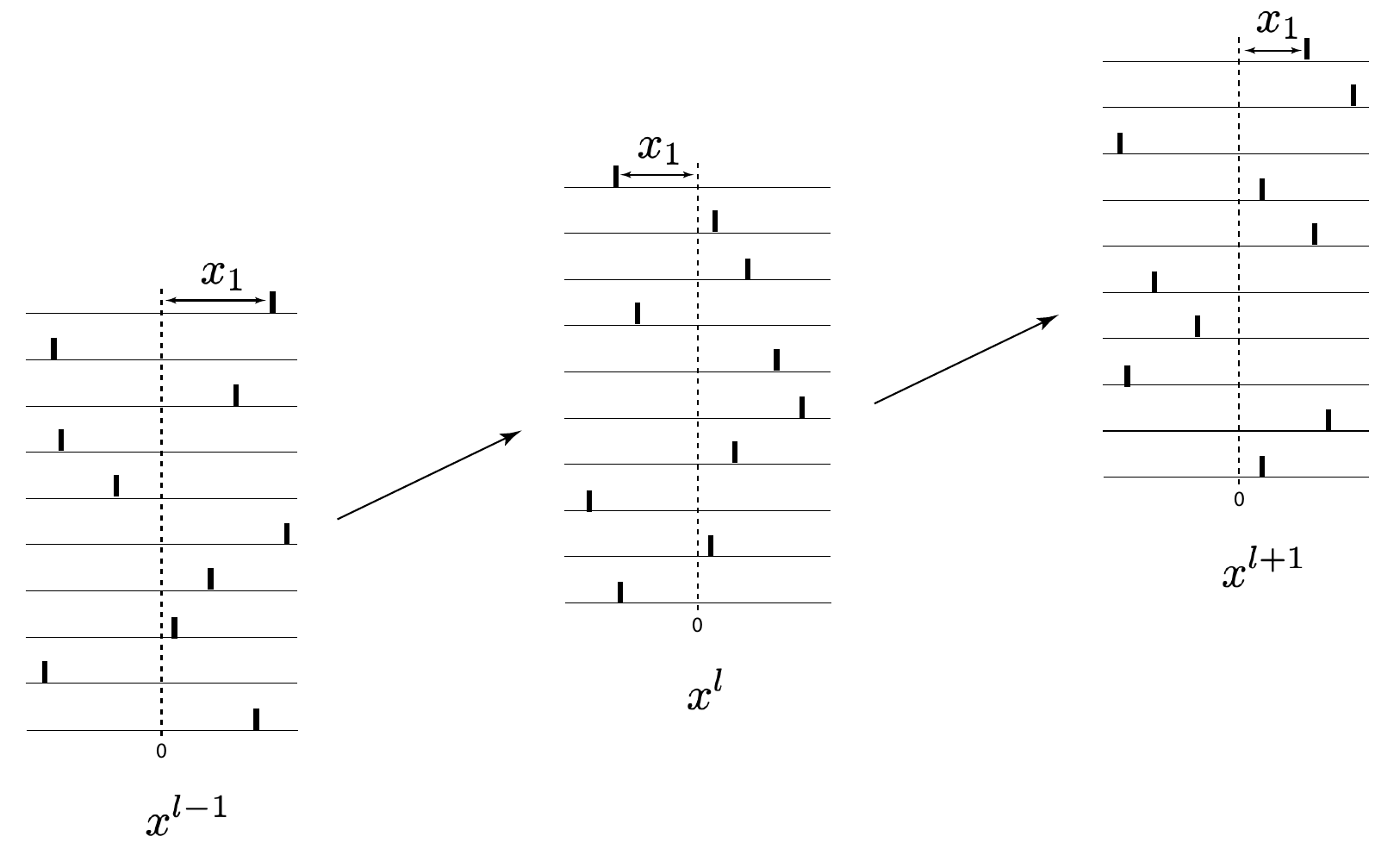}
    \caption{Spiking dynamics is treated as a transition from one spiking pattern $x^l$ to another spiking pattern $x^{l+1}$, where $l$ could denote different layers in a feedforward network or different cycles within the same layer. There is no time variable within each spiking pattern. Instead, it is described by the vector $x=(x_1,\ldots,x_n)\in\R^n$, where each $x_i$ is the latency (timing) of spike of the $i$th neuron relative to some time stamp $0$.}
    \label{L-1LL+1.pdf}
\end{figure}

The {\em first pillar} of this manifesto is to abstract away from individual neurons. We model the network transition as a direct map from latency vector to latency vector, i.e., from spiking patterns in one layer to spiking patterns in the next layer (or the same layer at the next periodic cycle), illustrated in Fig. \ref{L-1LL+1.pdf}. This is accomplished by detecting spiking patterns via look-up tables and using the tables to determine latencies in the next layer. Because of the continuity of latency vectors, we can differentiate them and use gradient descent for training. As a result, we can match the high performance of leading AI models, but implement them in a look-up table format that is inherently more efficient. 

The {\em second pillar} of this manifesto is the claim that spiking networks are nature's way of implementing look-up tables. Unlike matrix multiplication but similar to look-up tables, spiking networks store a vast number of parameters (synapses) yet use only a tiny fraction at any given time because most neurons are silent. Similarly to look-up tables, spiking networks determine the identity (index) of each fired neuron; and exactly like look-up tables, spiking networks utilize only the synapses from that fired neuron while all the other synapses remain untouched until needed.

The AI field has come a long way from the original view of neurons as gates in logical circuits (McCulloch-Pitts \cite{McPitts}) to the modern linear algebra view: neuronal states as vectors, transition from one state to another via matrix multiplication, and learning as adjustment of the weight matrices. It is time to take the analogous step and treat spiking patterns as the fundamental states of the network, the transition from one pattern to another via look-up tables, and learning as the adjustment of these look-up tables. This approach allows for the replacement of all matrix multiplication in deep networks and transformers with look-up tables, and it suggests a novel hardware architecture to accelerate these models, one that is distinct from GPUs and traditional neuromorphic systems \cite{neuromorphic}.

We discuss relative capacity of ANNs and SNNs (Sect. \ref{capacity of SNNs}), the relationship to locality-sensitive hashing (Sect. \ref{Appendix: LSH}), to mixture-of-experts (Sect. \ref{Section: MoE}), finite-state machines (Sect. \ref{Section: finite-state machines}), decision trees, random forests, and ferns (Sect. \ref{Section: forests and ferns}), spiking implementations of transformers via quantizations (Sect. \ref{Section: Transformer quantizations}), and neuromorphic systems (Sect. \ref{Section: neuromorphic}) in Appendix.

\section{The model}

To exploit the massive combinatorial capacity of spiking patterns -- the feature that gives our network its efficiency -- we need a way to distinguish different patterns, even those that differ by the timing of a single spike. For this, we consider a special case when all neurons fire exactly one spike during a certain interval, as in Fig. \ref{L-1LL+1.pdf}. Any such spiking pattern can be described by a vector of {\em latencies} $x\in\R^n$, where $n$ is the number of neurons. That is, the $i$th neuron fires a spike at a time $x_i$ relative to the center of the interval, so that $x_i$ could be positive or negative. In the case of feedforward networks, we want to describe the transition from spike latencies in the layer $x$ to the spike latencies in the next layer $y$. In the case of recurrent networks, we describe the transition from $x$ at one temporal cycle to the same $x$ at the next cycle.  

The transition from $x$ to $y$ in ANNs is done via multiplying $x$ by a matrix of synaptic weights and then applying a nonlinear function, such as ReLU. However, matrix multiplication distinguishes poorly between relative timings of individual spikes. A better way to detect spiking patterns is to compare pairwise latencies of neurons in $x$. This allows us to enumerate various patterns using discrete integer indices $j$. Then, we use these indices to select pre-stored synaptic vectors from look-up tables, which are then summed to form the output vector $y$, effectively translating one spiking pattern $x$ into another spiking pattern $y$ for the subsequent layer.

\begin{figure}[t]
    \centering
    \includegraphics[width=0.5\textwidth]{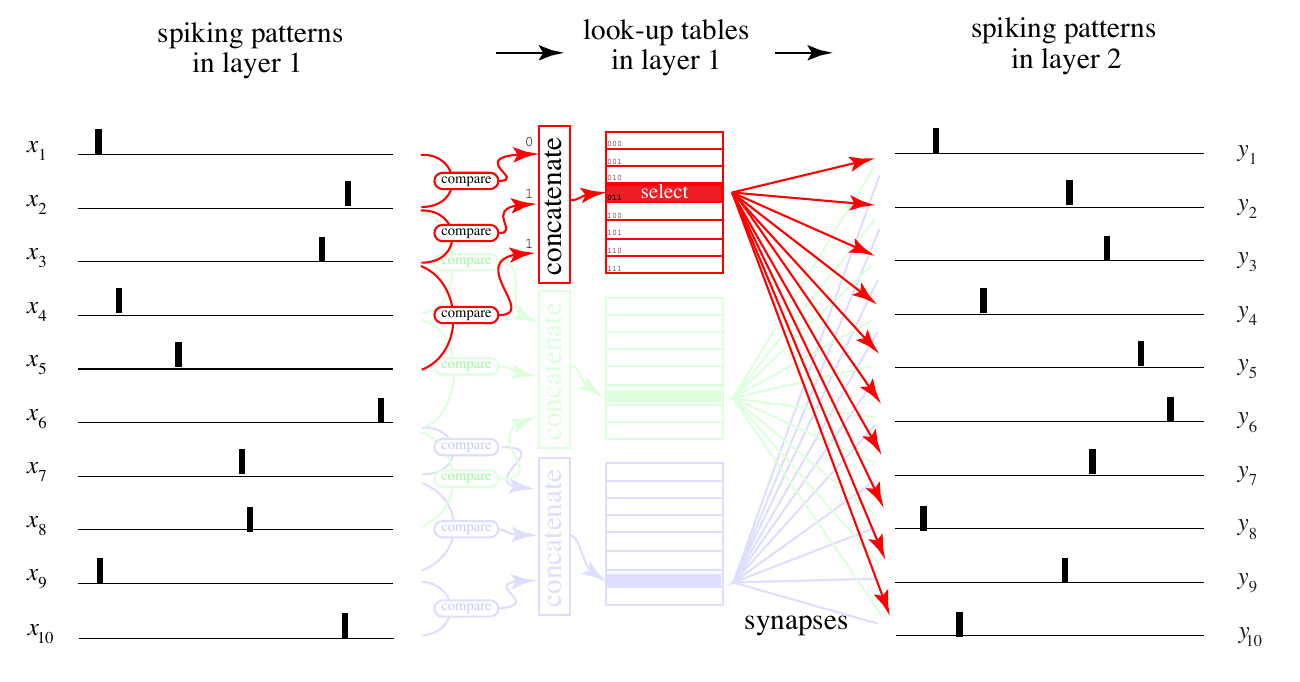}
    \caption{Various spiking patterns in layers 1 and 2 are represented by the vector of latencies (timings) $x, y\in\R^{10}$. Transition from $x$ to $y$ is done through $n_t=3$ look-up tables. Each look-up table monitors $n_c=3$ pairs of "anchor" neurons, selected randomly during initialization. The first (red) look-up table monitors $(x_1, x_2)$, $(x_2,x_3)$, and $(x_3,x_5)$. At each step, it evaluates 3 conditions, if $x_1 > x_2$, if $x_2>x_3$, and if $x_3>x_5$, each resulting in binary $0$ or $1$. Concatenating these 3 binary values results in an integer between $0$ and $7$, e.g., $011_{\rm binary} = 3$ in this figure (indexing starts with $000$). The integer is used as an index into the red table to retrieve the vector of synaptic values. These synaptic values are used as inputs to the vector of latencies $y$ in layer 2, resulting in various spiking patterns there.     }
    \label{polyb}
\end{figure}

Let us formalize this. The transformation from $x$ to $y$, illustrated in Fig. \ref{polyb}, involves $n_t$ look-up tables, each tracking $n_c$ "anchor" neurons using the following steps:
\begin{itemize}
\item {\sc Assign:} During initialization, assign $n_c$ random pairs of input neurons (anchors) to each table. 
\item {\sc Compare:} During inference, each table makes pairwise comparisons of latencies $x$ of its $n_c$ anchor  neurons.
\item {\sc Concatenate:} The resulting $n_c$ bits are concatenated to form a unique binary index $j$.
\item {\sc Select:} The index $j$ selects the row in the look-up table containing the vector of synaptic values to be used as an output from $x$ to the next layer $y$.  
\end{itemize}
For example, Fig. \ref{polyb} has $n_t=3$ tables, each performing $n_c=3$ comparisons, and hence having $2^{n_c}=8$ rows. Red table in the figure evaluates 3 conditions: $x_1>x_2$, \ $x_2>x_3$, and $x_3>x_5$, resulting in binary $011$ corresponding to the synaptic vector stored in row $j=3$ (row numbering starts with binary $000$). The entire row is used as the input for the next layer of neurons, $y$, achieving the desired transformation without any matrix multiplication. One can think of each look-up table as a sparse network of $2^{n_c}$ spiking "detector" neurons, each tuned to its own order of input spikes of the anchor neurons. Every spiking pattern excites a single detector neuron. 

\looseness=-1
Let vectors $a_i$ and $b_i$ denote the $n_c$ pairs of indices of the anchor neurons associated with the $i$th table. For example, the red table in the figure has $a_1 = (1, 2, 3)$ and $b_1=(2, 3, 5)$. The $i$th table looks at the pairs $u_{ir} = x_{a_{ir}} - x_{b_{ir}}$ and evaluates the condition $u_{ir} > 0$ to get $n_c$ binary values for all $r = 1, \ldots, n_c$. Then, the row index $j$ is just concatenation of these binary values
\begin{equation}
\label{j=H}
j = H_i(x) = \mbox{concat}( \ u_{i1}\!>\!0, \ \ldots, \ u_{in_c}\!>\!0 \ ), 
\end{equation}
which is a function of $x$. Each such index corresponds to a spiking pattern exhibited by anchor neurons with indices $a_i$ and $b_i$, and the table can distinguish $2^{n_c}$ such patterns. 

Let $s_{ijk}$ denote the synaptic value in table $i$, row $j$ (i.e., corresponding to polychronous pattern $j$), and column $k$ (i.e., projecting to $y_k$). Since selecting index $j$ is a function of $x$, we can write $s_{iH_i(x)k}$, but this notation is cumbersome. Instead, for simplicity, we will write $s_{ixk}$, keeping in mind that $x$ stands for an appropriate second index $j$ of $s_{ijk}$, different for different $i$. Now, the transition from $x$ to $y$ can be written as   
\begin{equation}
\label{sum s j}
y_k = \sum_{i=1}^{n_t} s_{ijk}  \ \ \mbox{ where } \ \ j=H_i(x)
\end{equation}
or, inserting $j=H_i(x)$ into the index, as 
\[
y_k = \sum_{i=1}^{n_t} s_{iH_i(x)k} 
\ \ \ \ \ \mbox{or } \ \ \ \
y_k = \sum_{i=1}^{n_t} s_{ixk} 
\]
or using vector notation $S_{ix} = (s_{ix1},\ldots,s_{ixn}) \in\R^n$ as
\begin{equation}
\label{y=Sx}
y = \sum_{i=1}^{n_t} S_{ix} \ \ \ \ \mbox{or } \ \ \ \ y = S_x\;
\end{equation}
All these notations are equivalent. $S_x$ is just a shorthand to denote the sum of constant vectors. Variable $x$ only plays a role in determining which rows from each look-up table $S_i$ to use; see the first half of the pseudocode in Fig. \ref{pseudocode}. 

Motivated by the dynamics of cortical neurons (see Sect. \ref{Appendix: Latency}), we use simple addition for the update from $x$ to $y$. Thus, it is a misnomer to call $s_{ijk}$ weights, since the term "weights" implies that they are multiplied by something, whereas in spiking networks the synapses are always added to something. The Eq. (\ref{y=Sx}) is nevertheless a nonlinear operation because of the selection of the index $j=H_i(x)$, i.e., selecting which $s_{ijk}$ to add. The combined action of summing the look-up tables is defined as the {\em look-up transformation} $S_x$ (LUT $S_x$). In the Appendix, we will discuss other possible choices of $H_i(x)$. 

We now make two observations to justify Fig. \ref{memory-processing}: 
\begin{itemize}
\item {\sc Low processing demand}. The conversion from a spiking pattern to a look-up table is just a bit concatenation. A look-up table with $n_c$ inputs distinguishes $2^{n_c}$ different  patterns at no additional processing cost. A LUT with $n_t$ look-up tables has the capacity to distinguish $2^{n_tn_c}$ different spiking patterns -- a giant number.
\item {\sc Low memory demand}. While memory footprint of each look-up table might be large, only one row is retrieved during each forward pass, resulting in low memory bandwidth demand. Memory storage is cheap;  memory access is expensive. 
\end{itemize} 
The representation of a practically infinite set of all possible spiking patterns as a set of sparse indices of look-up tables is a form of locality-sensitive hashing \cite{LSH}, and it is arguably the most efficient method of encoding and retrieving information. We explore this relationship in Sect. \ref{Appendix: LSH}.

\section{Gradients and error backpropagation}

Learning through gradient descent involves partial derivatives of (\ref{y=Sx}), which are discontinuous with respect to $x$. Indeed, $S_{ix}$ is piecewise constant unless $x_{a_{ir}}=x_{b_{ir}}$ for some $r$, in which case a small perturbation flips the order of spikes, changes the index $j = H_i(x)$, and flips $S_{ij}$ to some other value.  This complication is related to the general problem of applying error backpropagation to all SNNs -- small perturbations tweak timings of spikes, but if a neuron is near its threshold, perturbations can add or remove a spike, making the whole dynamics discontinuous.  However, due to the special way we represent spiking patterns, there is a path to do error backpropagation through a look-up table.

\begin{figure}[t]
    \centering
    \includegraphics[width=0.49\textwidth]{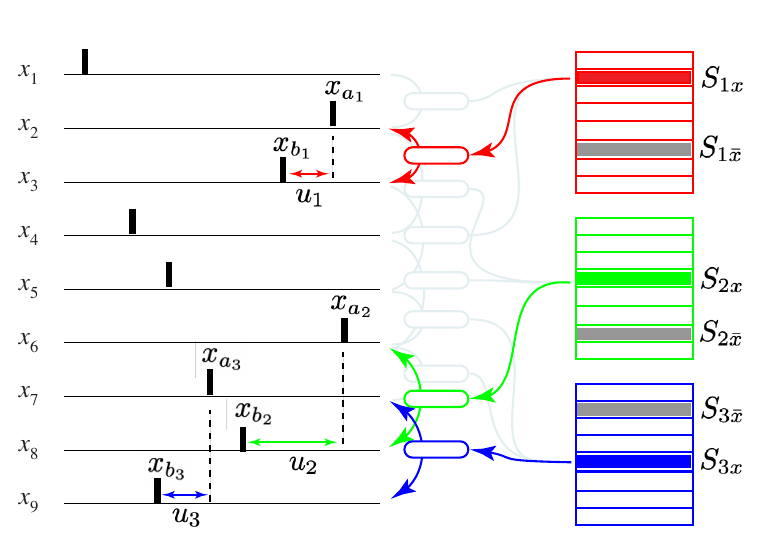}
    \caption{Error backpropagation through look-up tables: Selection of the minimal anchor pairs, $u_i$, for the learning rule (\ref{grad y xab}). Flipping the order of spikes in any such pair will result in a change in the index of the corresponding table. Notice that the smallest latency difference is between $x_7$ and $x_8$, but none of the look-up tables compares these two neurons and none will be affected by flipping their order. }
    \label{learning_explanation}
\end{figure}

\subsection{Surrogate gradient}

\begin{figure}[t]
    \centering
    \includegraphics[width=0.4\textwidth]{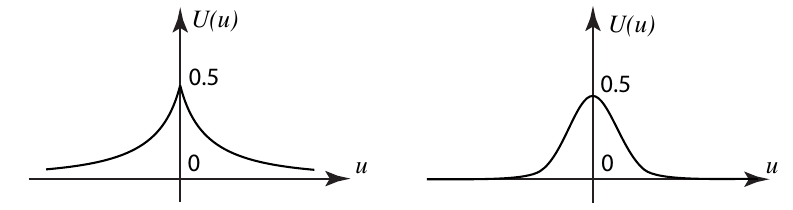}
    \caption{Examples of symmetrical U-shaped "uncertainty" functions to smoothen the transitions from one polychronous pattern to another.}
    \label{Ufunction}
\end{figure}

For each look-up table, let us denote the anchor pair having the smallest absolute value as $u_{i} = x_{a_{i}} - x_{b_{i}}$, as in Fig. \ref{learning_explanation}. Among all other latency differences monitored by table $i$, this $u_i$ requires the smallest perturbation of $x$ to flip its sign. This flipping corresponds to the index in the $i$th look-up table, $j=H_i(x)$, flipping one bit to become $\bar{j} = H_i(\bar{x})$, causing the synaptic vector $S_{ij}$ to flip to $S_{i\bar{j}}$, i.e., $S_{ix}$ to $S_{i\bar{x}}$. When $u_{i} \approx 0$, we are not sure whether we should use synaptic values $S_{ix}$ or flipped values $S_{i\bar{x}}$. And when $u_{i}=0$, probably the average of the two is the best choice. Let continuous symmetrical function $U(u)$ in Fig. \ref{Ufunction} represent our "uncertainty" about the choice. It should satisfy $U(u) \approx 0$ for large $|u|$ and $U(0) = 0.5$ reflecting the biggest uncertainty. 

Then, we can use $U$ to make the dynamics smooth, at least in the small neighborhood of $x$ that only includes flipping of signs of $u_i$.  This is similar to the standard method of "surrogate gradients" in spiking networks \cite{Neftci2019,spikingbackprop}.   We replace (\ref{y=Sx}) with a surrogate function
\begin{equation}
\label{surrogate}
y =  \sum_{i=1}^{n_t} \left\{ S_{ix}   + U(x_{a_{i}} - x_{b_{i}}) (S_{i\bar{x}} - S_{ix} )  \right\} \;. 
\end{equation}
Basically, as $u_{i} \rightarrow 0$, the synaptic values approach the average $(S_{ix}+S_{i\bar{x}})/2$. Other choices, e.g. applying $U$ to all pairs or ignoring the flipping, are discussed in Sect. \ref{Backprop without MatMul}. 

Differentiating this equation with respect to $x_m$ gives zero Jacobian vector $\partial y/\partial x_m=0$ for all $m$, except two values, $m={a_i}$ and $m={b_i}$. In this case   
\begin{equation}
\label{grad y xab}
\frac{\partial y \ }{\partial x_{a_{i}}} = - \frac{\partial y \ }{\partial x_{b_{i}}} = U'(x_{a_{i}} - x_{b_{i}}) (S_{i\bar{x}} - S_{ix})\;.
\end{equation}
Oppositely signed derivatives represent the two opposing forces that either pull $x_{a_i}$ and $x_{b_i}$ together or push them apart in Fig. \ref{learning_explanation}. When neuron $x_m$ anchors the minimal comparison across multiple look-up tables, the derivatives are summed.
  
We only use $U$ in the training mode to calculate gradients, but we use (\ref{y=Sx}) without $U$ in the inference mode. The choice of $U$ changes the training results very little. In simulations, we use $U(u) = 0.5/(1+ |u|)$ in Fig. \ref{Ufunction}, left. Notice that neither $S$ nor $x$ have any scale in the model. Using the nonlinear function $U$ introduces the scale, limiting the plasticity window to sufficiently small latency differences $u \approx 0$.

\subsection{Error backpropagation}

Let us consider a generic multi-layer SNN of the form 
\begin{equation}
\label{deepNN}
x^{l+1} = S^l_{x^l}\;,
\end{equation}
where $l$ is the layer number. Let $\L$ denote the loss (error) function that we want to minimize. Differentiating $\L$ with respect to each $x^l_m$ and applying the chain rule results in the standard recursive expression
\[
\frac{\partial \L \ }{\partial x^l_m} =   \frac{\partial \L \ \ }{\partial x^{l+1}} \cdot \frac{\ \partial x^{l+1} \ }{\partial x^l_m}
\]
that shows how the error at layer $l+1$ propagates back to layer $l$. Here, the Jacobian vectors $\partial x^{l+1} / \partial x^l_m$ are either $0$ or are given by the equation (\ref{grad y xab}) with $y = x^{l+1}$.
Introducing variable 
\begin{equation}
\label{gi}
g^l_{i} =   \frac{\partial \L \ \ }{\partial x^{l+1}} \cdot \left(S^l_{i\bar{x}^l} - S^l_{i{x}^l}\right)\;,
\end{equation}
we can write the gradient backpropagation equation as   
\begin{equation}
\label{dLdx}
\frac{\partial \L \ }{\partial x^l_{a^l_{i}}} = - \frac{\partial \L \ }{\partial x^l_{b^l_{i}}} = U'(x^l_{a^l_{i}} - x^l_{b^l_{i}}) \; g^l_{i}\;.
\end{equation}
This learning rule, used in all simulations in this paper, has a simple interpretation: $g_i$ measures which synaptic vector, $S_{i\bar{x}}$ or $S_{i{x}}$, is better aligned with the gradient vector $\partial \L/\partial x^{l+1}$. If $g_i$ is negative, then keeping the current $S_{ix}$ is better and the minimal latency difference should be further away from zero. If $g_i$ is positive, then flipping is better and the minimal difference should be closer to zero. This has to be done for all look-up tables in all layers. A geometrical illustration of this rule is in Fig. \ref{generalization}.

Each table contributes to the gradient two equal numbers of the opposite sign, $\partial \L/\partial x^l_{a}$ and $\partial \L/\partial x^l_{b}$. As a result, the gradient vector $\partial \L/\partial x^l$ has zero mean, which causes the synaptic vectors, $S^l_{ix}$, and hence the latency vectors, $x^l$, to maintain zero mean, providing self-regularization and stability.

In Appendix, Sect. \ref{Backprop without MatMul}, we explore variants of this rule. One variant takes into account all latency differences for all $n_c$ comparisons within each table. Another one considers only one table (with minimal $u_i$) in the entire layer to compute the gradient. That latter rule can be implemented without matrix multiplication, hinting on the possibility that the same spiking hardware can do both, training and inference.

\section{Examples}

To illustrate the versatility of our approach, we demonstrate the conversion of deep ANNs, recurrent networks (RNNs), and Transformers into the SNN form. We adhere to a single rule: preserve the model's overall architecture, treat each activation vector $x$ as a latency vector, and simply replace all matrix multiplications with look-up table transformations. This process is akin to compiling high-level code to specialized assembly language, where the fundamental complexity is managed beneath the surface.

\begin{figure}[t]
    \centering
    \includegraphics[width=0.44\textwidth]{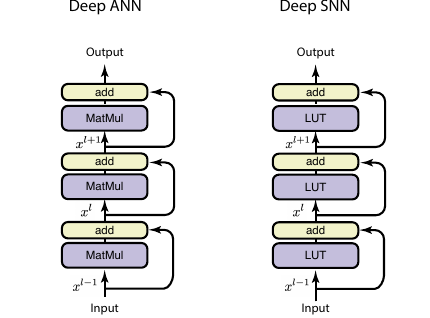}
    \caption{Replacing matrix multiplication with look-up transformation (\ref{y=Sx}) converts a deep ANN into a deep SNN.}
    \label{DeepANNSNN.pdf}
\end{figure}

\subsection{Deep SNN}
\label{Deep Learning}

Consider the simplest ANN -- multi-layer perceptron (MLP) with residual (skip \cite{resnet}) connections
\begin{equation}
\label{deepANN}
x^{l+1} = x^l + \mbox{ReLU}( M^lx^l + b^l)\;.
\end{equation}
Here synaptic weight matrix $M^l\in\R^{n \times n}$ and bias vector $b^l\in\R^n$ are layer dependent, $l=0,\ldots,L$, and nonlinearity is conveyed via ReLU$(u) = $ max$(0, u)$. This is a fundamental block in many ANN models, including the feedforward layer (FFN) in transformer architecture.

\begin{figure}[!htb] 
\centering
\caption{Pseudocode of the deep SNN (\ref{DeepSNN}).}
\begin{minipage}{\columnwidth}
\small
\vspace*{0.2cm}
\begin{algorithmic}[1]
    \State \textbf{Initialize look-up tables}: zero or random synaptic vectors $S^l_{ij}$ 
    \State \textbf{Initialize anchors}: random pairs $a^l_{ir} \neq b^l_{ir}$ for each table 
    \vspace{0.15cm}
    \State \textbf{Forward pass (inference):} 
    \State $x^0 \gets $ input
    \For{$l = 0 \textbf{ to } L$}    \, \ \ \ \ \ \ \ \ \ \ \ \ \ \, \ \ \ //  all layers
        \State $x^{l+1} \gets x^l$    \, \ \ \ \ \ \ \ \ \ \ \ \ \ \ \ \ \ \ \ \ \ \ // residual connections  
        \For{$i = 1 \textbf{ to } n_t$}  \ \ \ \ \  \ \ \ \ \ \ \ \ \ //  all look-up tables 
            \State $j \gets 0$ \, \ \ \ \ \ \ \ \ \ \ \ \ \ \ \ \ \ \ \ \ \ \ \ // look-up table index
            \For{$r = 1 \textbf{ to } n_c$}  \ \ \ \ \ \ \ \  \ //  all comparisons 
               \State $u_{ir} \gets x_{a_{ir}} - x_{b_{ir}}$ \ \ \ \ \ // anchor neurons
               \State $j \gets \textbf{concat}(j, u_{ir} > 0) $  \negthinspace // Eq. \ref{j=H}: determine the index
            \EndFor 
            \State \textbf{cache} $j$ and $u_i, r_i$ {\bf corresponding to} $\min(|u_{ir}|)$
            \label{line:cache}
            \State $x^{l+1} \gets x^{l+1} +  S^l_{ij}$ \ \ \ \ \ \ \ \ \ \ \ \ \,  // Eq. \ref{deepNN}: synaptic transmission
        \EndFor 
    \EndFor 
    \State output $\gets x^{L+1}$ 
    \vspace{0.15cm}  
    \State \textbf{Backward pass (learning):} 
    \State $\mathcal{L} \gets$ CalculateLoss( output )
    \State $v^{L+1} \gets {\partial \mathcal{L}}/{\partial x^{L+1}}$  \ \ \ \ \ \ \ \ \ \ \ \ \ \ \ \ // last layer gradient 
    \For{$l = L \textbf{ down to } 0$}  \,  \ \ \ \ \ \ \ \ \, //  all layers backwards
        \State $v^l \gets v^{l+1}$ \ \ \ \ \ \ \ \ \ \ \ \ \ \ \ \ \ \ \ \ \ \ \, // residual connections
        \For{$i = 1 \textbf{ to } n_t$}  \ \ \ \  \ \ \ \ \ \ \ \ \, //  all look-up tables 
            \State $\textbf{uncache } j$ and $u_{i}, r_{i}$  \ \ \ \ \ \; // corresp. to $\min(|u_{ir}|)$
            \State $\bar{j} \gets j \text{ \bf xor } 2^{r_i}$  \ \ \ \ \ \ \ \ \ \ \ \ \ \,  // flip $r$th bit of $j$
             \vspace{0.05cm}
	    \State $g_{i} \gets v^{l+1} \cdot (S^l_{i\bar{j}} - S^l_{ij}) \,$ \ \ \  // Eq. \ref{gi}: gradient alignment
            \vspace{0.02cm}
            \State $v^l_a \gets v^l_a +  U'(u_{i}) \ g_{i}$ \ \ \ \ \ \ // Eq. \ref{dLdx} with $a=a^l_{ir_i}$
            \vspace{0.07cm}
            \State $v^l_b \gets v^l_b \, -  U'(u_{i}) \ g_{i}$ \ \ \ \  \ \ // Eq. \ref{dLdx} with $b=b^l_{ir_i}$ 
            \vspace{0.07cm}
            \State $S^l_{ij} \gets S^l_{ij} - \varepsilon v^{l+1} $ \ \ \ \ \ \ \ \; // synaptic update (learning)
            \vspace{0.05cm}
        \EndFor 
    \EndFor 
\end{algorithmic}
\end{minipage}
\label{pseudocode} 
\end{figure}

Figure \ref{DeepANNSNN.pdf} illustrates our general principle: The overall architectural flow of vectors $x^l\in\R^n$ is preserved; we simply replace the standard ReLU and MatMul operations with look-up table transformations:
\begin{equation}
\label{DeepSNN}
x^{l+1} = x^l + S^l_{x^l}\;.
\end{equation}
Notice in the pseudocode in Fig. \ref{pseudocode} that we do not need to keep activation vectors $x^l$ for backward (learning) pass. Instead, we rely on index caching (line \ref{line:cache}). This index caching becomes crucial for the V-index caching in the efficient SNN implementation of attention in Sect. \ref{Section: Attention} -- the SNN's equivalent of KV-caching in ANN transformers.

\subsection{Spiking RNN}

Recurrent neural networks (RNNs)  serve as the foundation of many impactful early LLMs such as Long Short-Term Memory (LSTM) \cite{LSTM, LSTM2} or Gated Recurrent Unit (GRU) \cite{GRU}. At any given time step $t$, the state of the network, $h_t$, is determined by the input $z_t$ and the previous, "hidden", state $h_{t-1}$ via some transformation $h_t = f(h_{t-1}, z_t)$ that typically involves matrix multiplications. The simplest implementation of an RNN is an Elman network \cite{Elman}:
 \begin{equation}
\label{eq RNN}
h_t = S_{h_{t-1}}+z_t\;, \ \ \ h_0=0\;,
\end{equation}
where LUT $S_h$ has the form (\ref{y=Sx}). 
In addition, it also has a learnable look-up table $E$ that embeds the $t$th input character into $z_t\in\R^n$ and a learnable LUT $U_{h_t}$ of the form (\ref{y=Sx}) that unembeds the hidden state $h_t$ into the output probability distribution over the vocabulary set. In Sect. \ref{Section: Spiking RNNs} we discuss other variants of spiking RNNs.

\subsection{SNN Transformer}
\label{Section: Attention}

\begin{figure}[t]
    \centering
    \includegraphics[width=0.35\textwidth]{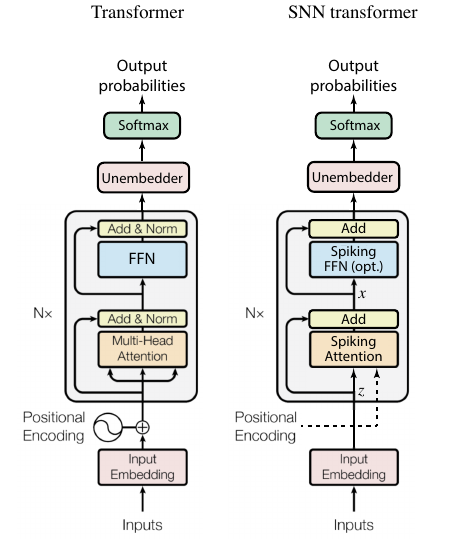}
    \caption{(a) Replacing an ANN transformer with SNN transformer. Each (optional) Feed Forward network is a single LUT (\ref{z=Sx}) and each attention head is the LUT (\ref{x=Vz}) depicted in Fig. \ref{attention.pdf}. }
    \label{transformer.pdf}
\end{figure}

Now, we are ready to implement an SNN transformer. The first step, common to both ANNs and SNNs and depicted in Fig. \ref{transformer.pdf}, is to use an embedder look-up table to convert each input token into an embedding vector $z_i\in\R^n$, where $i$ denotes the position of the token in the input string of length $n_{inp}$. The reverse is done at the last step, projecting an embedding vector back to vocabulary space and then taking the softmax to get probability distribution of the output tokens. ANN transformer un-embeds via matrix multiplication while SNN transformer uses LUT of the form (\ref{y=Sx}). Both, embedder and un-embedder are learnable.  The middle of the transformer is the "attention + feedforward" block repeated $N$ times. We denote inputs to the attention module as $z_i$ and outputs as $x_i$; the feedforward network (FFN) converts $x_i$ back to $z_i$. 

The SNN implementation of FFN layer,  
\begin{equation}
\label{z=Sx}
z_i = x_i + S_{x_i}\,,
\end{equation}
is just a single-layer LUT (\ref{DeepSNN}) with pseudocode in Fig. \ref{pseudocode}. We show later that this module is optional, owing to the inherent capabilities of SNN attention module.

The standard dot-product attention module,  
\begin{equation}
\label{attention}
\mbox{softmax}(\frac{\ QK^\top}{\sqrt{d_k}})\, V\;,
\end{equation}
has a built-in multiplication of matrices, so obtaining $Q$, $K$, and $V$ via LUTs would still leave softmax bottleneck and matrix multiplication of $V$ by the softmax attention scores. 

\begin{figure}[t]
    \centering
    \includegraphics[width=0.37\textwidth]{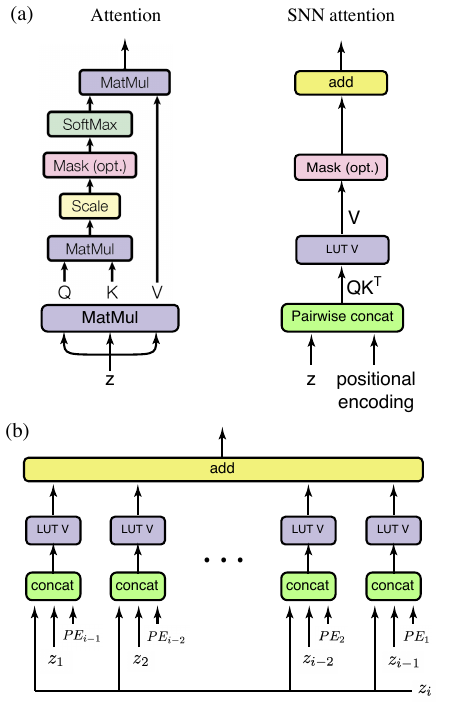}
    \caption{(a) Replacing a single head of transformer's attention (\ref{attention}) with SNN (\ref{x=Vz}).  (b) Zoom-in on the concatenation and look-up transformation (LUT). Optional mask is omitted for simplicity.}
    \label{attention.pdf}
\end{figure}

Among many approaches discussed in Appendix, Sect. \ref{Appendix: Attention}, the most efficient one is presented in Fig. \ref{attention.pdf}. It captures the long-distance dependencies between different input embeddings by concatenating input vectors
\begin{equation}
\label{concatenated z}
[z_i, z_j, PE_{i-j}] \in \R^{2n+p}
\end{equation}
for every Query-Key pair $z_i$ and $z_j$. Here, the relative positional encoder $PE_{i-j} \in \R^p$ is a learnable vector that encodes the relative distance from $z_i$ to $z_j$ \cite{Shaw}. If $p!>n_{inp}$, then $PE$ provides enough context to encode all relative positions as a pattern of spike latencies. 

Each such concatenated vector is then used as an input to  
\begin{equation}
\label{x=Vz}
x_i = z_i +  \sum_{j=1}^{i-1} V_{[z_i, z_j, PE_{i-j}]}\;,
\end{equation}
where LUT $V$ is the same for all embedding pairs, but different for different heads and layers. In self-attention, $z_i$ can only look at the earlier tokens, so we only consider $j<i$. Interestingly, including the "self" term $j=i$ into the sum reduced the performance of the model. Outputs of multiple attention heads are added without any weighting, though one can treat each look-up table as its own attention head. 

Notice that there is no softmax operation here. The attention network learns the relative magnitudes (attention scores) of value vectors $V_{[z_i, z_j, PE_{i-j}]}$ by being exposed to various pairs of embeddings at various locations. This is conceptually similar to how different attention heads learn different long-distance dependencies even though there is no direct competition between them. In Appendix, we also implemented the "QK transformation" via LUTs to get attention scores $a_{ij}=A_{[z_i, z_j, PE_{i-j}]}$, applied softmax, and then used them to scale up the value vectors $V$. We saw improved performance, which is likely due to the apparent increase in the number of trainable parameters. However, this version of SNN attention had matrix multiplication and softmax bottleneck, and hence was undesirable.

\subsection{V-index cache}

A naive approach to implement the SNN attention mechanism is to create $n_{inp} \times n_{inp}$ concatenated pairs of the form (\ref{concatenated z}) and then apply the LUT (\ref{x=Vz}) to each of them, resulting in quadratic scaling, $\mathcal{O}(n^2_{inp})$.

A better approach, one that achieves linear scaling, is as follows. For simplicity, assume that $V_z$ has only one look-up table, so we omit the subscript $i$. We apply the function $H$ (Eq. \ref{j=H}), using $n_c$ anchor neurons, to every embedding $z_{\text{pos}}$ to get the look-up row index $j_{\text{pos}}$. We similarly apply $H$, using a different set of $p$ anchor neurons, to every positional encoder $PE_{\text{pos}}$ to get $j^p_{\text{pos}}$. This pre-computation is performed once for all $n_{inp}$ positions. We call it V-index caching. Now, to obtain the index for any concatenated vector $[z_{\text{pos}_1}, z_{\text{pos}_2}, PE_{\text{pos}_1-\text{pos}_2}]$ needed by the look-up table $V_z$, we simply concatenate the three cached indices: $j_{\text{pos}_1}, j_{\text{pos}_2}$, and $j^p_{\text{pos}_1-\text{pos}_2}$. This mechanism is equivalent to the one in Fig. \ref{attention.pdf}b with $z_i$, $z_j$, and $PE_{i-j}$ being replaced by their respective cached indices. 
 
 V-index cache suggests a simple interpretation of how our approach relates to the standard attention mechanism (\ref{attention}): computing an index for $z_i$ corresponds to computing the query vector $Q_i$. Computing the index for $z_j$ corresponds to computing the key vector $K_j$. Concatenating them corresponds to the dot-product $Q_iK^\top_j$, and using the concatenated index in the LUT (\ref{x=Vz}) corresponds to multiplying the dot product by the value vector $V_j$.

\section{Experiments}
\label{Results}
\label{Experiments}

There should be no expectation that SNNs will immediately outperform ANNs because ANNs have benefited from two decades of technological refinement, yielding performance-boosting techniques such as dropout, L2 regularization, or Adam optimization. All these performance-boosting techniques have yet to be invented for SNNs of the form (\ref{y=Sx}).

In contrast, it is reasonable to expect that SNNs will outperform ANNs when compared apples-to-apples—that is, without performance-improving regularization or optimization. To illustrate this point, we use the standard byte-size token prediction task \cite{Evaluation of NLPs} -- providing the model with 32-character snippets of English text and requiring it to predict the next character. We performed no hyperparameter search, relying solely on sanity check with a few initial choices of $n$, $n_t$, and $n_c$, and initialize all synapses to $0$. All SNN simulations were performed on author's 2022 MacBook Air. Simulations of SNN transformer were stopped after its validation perplexity dropped below 2.0 (bits-per-character dropped below 1.00). The code is available at https://github.com/izhikevich/SNN.

\subsection{Spiking RNN}

Let us illustrate a few interesting properties of SNNs using spiking RNN of the form (\ref{eq RNN}), with parameters detailed in Table \ref{RNN table}. 

\begin{table}[ht]
\centering
\caption{Parameter values of spiking RNNs (\ref{eq RNN})}
\begin{tabular}{lcc}
\toprule
 & Parameter & Value \\
\midrule
Context size & $n_{inp}$ & 32 \\
Dimension of hidden state $h_t$ & $n$ \ & 64 \\
Memory footprint of embedder table $E$ &  $256\times n$ & 16K  \\
\midrule
\multicolumn{3}{l}{\bf look-up transformation $S_h$} \\
Number of look-up tables  & $n_t$ & 64 \\
Number of comparisons in each table & $n_c$ & 10 \\
Memory footprint & $n_t \, 2^{n_c}n$ &  4M \\
Memory bandwidth per token & $2n_tn_c + n_tn$ &  5.4K \\
\midrule
\multicolumn{3}{l}{\bf unembedder look-up transformation $U_h$} \\
Number of look-up tables  & $n_t$ & 64 \\
Number of comparisons in each table & $n_c$ & 6 \\
Memory footprint & $n_t \, 2^{n_c} 256$ &  1M \\
Memory bandwidth per token & $2n_tn_c + n_t256$ &  17K \\
\midrule
\bf Size of the model &  &  \bf 5M \\
\bottomrule
\end{tabular}
\label{RNN table}
\end{table}

It might seem counterintuitive in the context of ANNs that the memory bandwidth of the spiking RNN is significantly smaller than its synaptic footprint, since the two are typically the same in ANNs. Let us perform a careful calculation: for each input, $S_h$ needs to load $2n_tn_c$ anchor neurons to determine the indices $j$, and then load $n_t$ synaptic vectors $S_{ij}\in\R^n$, requiring only 5.4K values. The rest of the 4M synaptic values are entirely passive, becoming relevant only when a new input triggers the generation of different indices $j$. This structure enables efficient scaling: increasing the number of comparisons, $n_c$, by 1 doubles the synaptic memory footprint, yet only marginally increases the active computation (FLOPs or memory bandwidth). This unique feature allows SNNs to learn more data without substantially increasing the computational cost, thereby boosting the data-per-FLOP and data-per-bandwidth efficiency ratios.

A notable characteristic of spiking RNNs is their resilience to vanishing or exploding gradients, an advantage that stems from two main architectural features. First, the LUT $S_{h_t}$ selects different rows $j$ for distinct hidden states $h_t$ during a forward pass, thereby avoiding the repetitive multiplication by the same synaptic matrix that characterizes standard RNNs. Second, although small gradients may lead to small weights and correspondingly small magnitude of hidden state vectors, the selection of the look-up index is based on pairwise comparisons of elements of these vectors. The network’s function is largely unaffected by the absolute magnitude of the components; whether they all have a magnitude of $10^{+10}$ or $10^{-10}$ does not alter the outcome of pairwise comparisons and hence the output to the next layer. The magnitudes, however, affect the learning rate via the uncertainty function $U'$ in (\ref{dLdx}). 
  
The standard metric for measuring byte-size LLM performance is Bits-Per-Character (BPC), which is the cross-entropy loss calculated using the base-2 logarithm rather than the common natural logarithm. As Table \ref{BPC table} shows, despite its small size and the absence of LSTM gates, regularization, or optimizations, the 5M-parameter spiking RNN would have been competitive with the state-of-the-art  LSTMs prior to the invention of Transformers in 2017.

 \begin{table}[ht]
\centering
\caption{Entropy in bits-per-character (BPC)}
\begin{tabular}{lcrc}
\toprule
Model & year & size & BPC \\
\midrule
MI-LSTM \cite{Wu2016} & 2016 & 17M & 1.44 \\
mLSTM \cite{Krauser} &  2016 & 10M & 1.40 \\
\bf Spiking RNN (Eq. \ref{eq RNN}) & &  \bf 5M & \bf 1.39 \\
BN LSTM \cite{cooijmans}  &  2016 & 16M & 1.36 \\
HM-LSTM \cite{Chung2016} &  2016 & 35M & 1.32 \\
LN HM-LSTM \cite{Chung} &   2016 & 35M & 1.29 \\
\bf SNN transformer (Eq. \ref{z=Sx}, \ref{x=Vz}) &  &  \bf 806M & \bf 0.99 \\
\bottomrule
\end{tabular}
\label{BPC table}
\end{table}

\subsection{SNN Transformers}
\label{Results: SNN Transformers}

We use the same byte-size token prediction task as above to compare the performance of ANN and SNN decoder-only (GPT-like) transformers in Fig. \ref{transformer.pdf}. For the ANN transformer, we use exactly the same parameters as in the original paper \cite{transformer}, i.e., $N=6$ layers, $h=8$ heads per layer, $d_{model}=512$, $d_{ff}=2048$, $d_k=64$.  The ANN simulation, provided by Taylor Kergan from UCSC, was performed without auxiliary training, such as dropouts, Adam optimization, or RMSNorm. Parameters of SNN transformers are provided in Table \ref{SNN transformer parameters}.  

\begin{table}[ht]
\centering
\caption{Parameter values of SNN transformers in Fig. \ref{transformer_sim.pdf}}
\begin{tabular}{lccc}
\toprule
 & Parameter & SNN & Attention-only \\
 & & \hspace{-0.2cm} (green curve) \hspace{-0.2cm} & \hspace{-0.2cm} (blue curve) \hspace{-0.2cm} \\
\midrule
Context size & $n_{inp}$ & 32 & 32 \\
Layers & $N$ & 6 & 6 \\
Embedding dimension & $n$ & 32 & 16 \\
\midrule
\multicolumn{3}{l}{\bf Attention $V_z$ } \\
Heads & $h$ & 4 & 1 \\
Look-up tables & $n_t$ & 16 & 10 \\
Comparisons & $n_c$ & 6 & 6 \\
Positional dimension & $p$ & 4 & 4 \\
Memory footprint &  $n_t\, 2^{2n_c+p} n$ & 33M &  10.5M \\
\midrule
\multicolumn{2}{l}{\bf Feedforward network $S_x$} & & removed \\
Look-up tables & $n_t$ & 16 & - \\
Comparisons & $n_c$ & 6 & - \\
Memory footprint &  $n_t\, 2^{n_c} n$ & 33K &  - \\
\midrule
\bf Size of the model & & \bf 806M & \bf 63M \\
\bottomrule
\end{tabular}
\label{SNN transformer parameters}
\end{table}

Notice the striking difference in memory allocation between ANN and SNN transformers. In ANNs, the majority of the synaptic memory is consumed by the feedforward network. Conversely, most synaptic memory in SNN transformers is consumed by the attention layer. This allocation is driven by the large size of the pairwise concatenated vectors (in Eq. \ref{concatenated z}), which necessitates many comparisons and exponentially increases the size of each look-up table to $2^{2n_c+p}$. This large memory footprint is the price paid for the greater efficiency of attention in SNN transformers. 

We use the parameters of the attention-only SNN transformer to compare resource demands of the ANN and SNN implementations in Table \ref{Table: Transformer Resources}. Notice the 10,000-fold difference in memory bandwidth demand:  ANNs need to load all $W$ matrices for each forward pass. Batch learning mitigates this issue by spreading the loading cost over many training examples, thereby narrowing the gap with SNNs.
 
\begin{table}[ht]
\scriptsize
\centering
\caption{Resource Demand for Inference per Layer per head in Fig. \ref{transformer.pdf}}
\begin{tabular}{lcc}
\toprule
 & \textbf{Transformer} & \textbf{SNN transformer} \\
\toprule
\multicolumn{3}{l}{Computational cost} \\
\midrule
{$QK^\top$ multiplications} & \hspace{-0.9cm} $2d_kn^2_{inp}+2d^2_{model}n_{inp}$ \hspace{-0.9cm} & 0  \\
{$QK^\top$ summations} & \hspace{-0.1cm} $2d_kn^2_{inp}+2d^2_{model}n_{inp}$ & 0 \\
$V$ and $O$ multiplications  \hspace{-0.8cm} & \hspace{-0.9cm} $2d_kn^2_{inp}+4d^2_{model}n_{inp}$ \hspace{-0.9cm} & 0 \\
$V$ and $O$ summations \hspace{-0.9cm} & \hspace{-0.9cm} $2d_kn^2_{inp}+4d^2_{model}n_{inp}$ \hspace{-0.9cm} & $n_tnn^2_{inp}$ \\
FFN multiplications & $8d^2_{model}n_{inp}$ & 0 \\
FFN summations & $8d^2_{model}n_{inp}$ & $n_tnn_{inp}$ \\
{Comparisons \& concatenations} \hspace{-0.5cm} & 0 & $2n_tn_cn_{inp}$  \\
\bf {Total } & \bf 235,405,312 & \bf 172,800  \\
\toprule
\multicolumn{3}{l}{{Memory footprint}} \\
\midrule
{$W^Q$, $W^K$, $W^V$, $W^O$ matrices } \hspace{-1cm} & $4d^2_{model}$ &   \\
{FFN's $W^1$ and $W^2$ matrices } \hspace{-1cm} & $8d^2_{model}$ &   \\
{$V_x$ look-up transformation } \hspace{-0.5cm} &  & $n_tn2^{2n_c+p}$  \\
{$S_x$ look-up transformation } \hspace{-0.5cm} &  & $n_tn2^{n_c}$  \\
\bf {Total } \hspace{-0.5cm} & \bf 3,145,728 &  \bf 10,496,000 \\
\toprule
\multicolumn{3}{l}{{Attention memory bandwidth demand per new token}} \\
\midrule
{$W^Q$, $W^K$, $W^V$, $W^O$ matrices } \hspace{-1cm} & $4d^2_{model}$ &   \\
{KV-cache }  & $(d_k+d_{model})n_{inp}$ &   \\
{Anchor neurons } &  & $2n_tn_c$  \\
{V-index cache} &  & $3n_tn_{inp}$  \\
\bf {Total  } \hspace{-0.5cm} & \bf 1,048,576+576$n_{inp}$ & \bf 120+30$n_{inp}$  \\
\bottomrule
\end{tabular}
\label{Table: Transformer Resources}
\end{table}

\begin{figure}[t]
    \centering
    \includegraphics[width=0.34\textwidth]{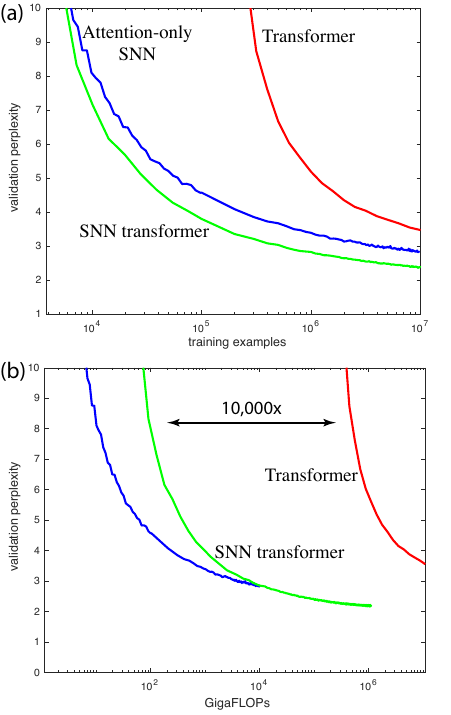}
    \caption{ Illustration of (a) the 50-fold difference in learning convergence rates and (b) the 10,000-fold difference in required computational resources of ANN and SNN transformers. Training was performed using the Adam learning rate scheduler that was optimized for ANN performance \cite{transformer}. Notice that the attention-only SNN has 16-dimension embedding space and a single attention head, whereas the ANN transformer has 512-dimensional embedding space and 8 heads.}
    \label{transformer_sim.pdf}
\end{figure}

Figure \ref{transformer_sim.pdf} illustrates another striking difference between ANN and SNN transformers: the learning convergence rate, as measured by performance on a validation dataset during training. This almost 50-fold difference cannot be dismissed as being due to a wrong learning rate parameter, especially since we used the Adam learning rate scheduler with parameters optimized for the ANN transformer, not the SNN one.

{\sc Ablation study}. Since SNN transformers are built on an architecture fundamentally optimized for efficiency and low memory bandwidth demand, we experimented with reducing the embedding dimension to $n=16$. This configuration still has greater encoding capacity, $16!$, than the linear encoding capacity of $\R^{512}$ -- the embedding dimension of ANN transformer, discussed in Sect. \ref{Memory representations}. To further test the benefits of SNNs, we remove the feedforward layer completely. In this "attention-only" transformer, plotted as blue curve in the figure, the output from attention module in layer $l$ goes directly to the attention module in layer $l+1$ and the whole architecture is just a stack of $N=6$ blocks in Fig. \ref{attention.pdf}b. Unlike scaled dot-product attention (\ref{attention}) of the ANN transformer, LUTs are inherently nonlinear. Therefore, the SNN attention is all you need to perform both token mixing and channel mixing.

\begin{figure}[t]
    \centering
    \includegraphics[width=0.46\textwidth]{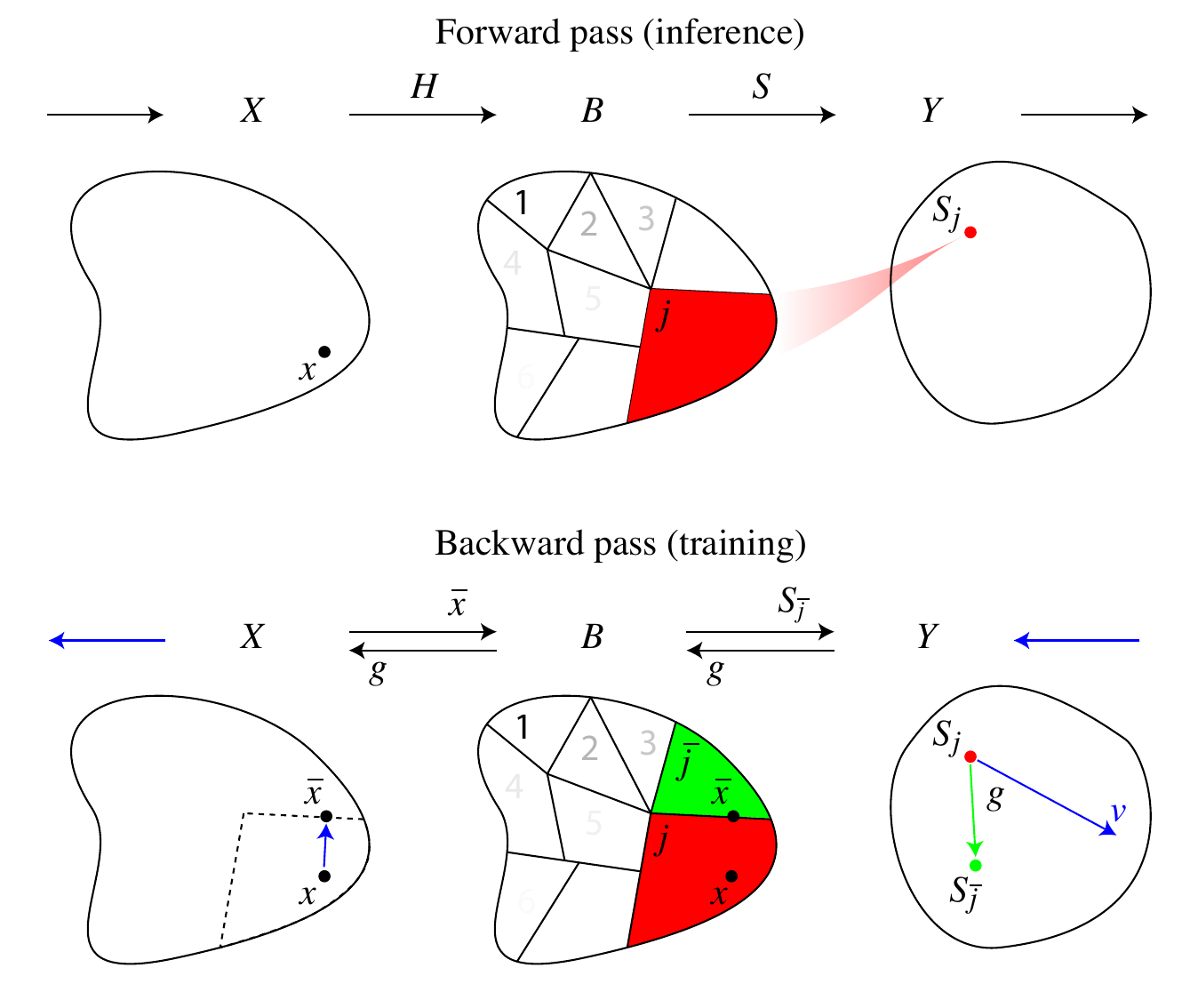}
    \caption{A generalization of the SNN framework presented in this manifesto: {\em Forward pass} of state $x$: A locality-sensitive hash function $H$ partitions the input space $X$ into a finite set of buckets $B$, numbered 1, 2, 3, etc. An input $x \in X$ is mapped to the bucket with index $j=H(x)$, which projects to a point $S_j \in Y$ in the output space. Consequently, all points within the red bucket project to the same output $S_j$, promoting generalization during inference.  {\em Backward pass} of blue vector $v$: Find the nearest neighbor $\bar{x}$ to $x$ that causes the hash index to flip, resulting in a different (green) bucket $\bar{j}=H(\bar{x})$. This transition moves the output from $S_j$ to $S_{\bar{j}}$. We measure the alignment of the resulting vector shift ($S_{\bar{j}}-S_j$) with the vector $v$ using a scalar $g$ (e.g., via dot product or projection on $v$). The surrogate (back-propagated) image of $v$ is along the vector $\bar{x}-x$ scaled by $g$ and function $U'(|\bar{x}-x|)$ defined in Fig. \ref{Ufunction}. If $g<0$, the surrogate vector points away from $\bar{x}$.  }
    \label{generalization}
\end{figure}

\section{Spiking as Locality-Sensitive Hashing (LSH)}
\label{Appendix: LSH}

The function (\ref{j=H}) that maps latency vectors $x\in\R^n$ to integer indices $j$ has a property of LSH (see \cite{LSH}): nearby vectors $x$ are mapped to the same index with high probability, whereas distant vectors are not.   
The LSH perspective inspires the generalized framework illustrated in Fig. \ref{generalization}.
\begin{itemize}
\item {\em Forward pass (the hash)}: The process of partitioning the continuous input space $X$ into a huge but finite number of discrete, non-overlapping hash buckets B using the classic definition of LSH. Each spiking pattern falls into a hash bucket. Small perturbations of input fall into the same bucket, reducing sensitivity to noise, decreasing overfitting, and promoting the generalization in the network.  
\item {\em Backward pass (the gradient)}: Since the hash function is not differentiable, learning relies on surrogate gradients, which are computed by determining the nearest neighbor, $\bar{x}$, of $x$ belonging to a different bucket, $\bar{j}$. Then, the surrogate gradient has the direction of the difference vector, $\bar{x}-x$, scaled by the error alignment $g$. 
\end{itemize}
Our implementation of SNN partitions $X=\R^n$ into $2^{n_tn_c}$ discrete buckets. The output of each bucket, $S_j$ is defined by (\ref{y=Sx}). Learning is just a nudge of $S_j$ along vector $v$. Error backpropagation is achieved by identifying the minimal input perturbation $\bar{x}$ that flips the relative spike order of a single anchor pair, $u_i$, in each table $i$. The resulting output shift ($S_{i\bar{j}}-S_{ij}$) is used in (\ref{gi}) to measure the alignment, $g_i$, with the error vector $v$. The surrogate gradient (\ref{dLdx}) is the average of all such perturbations $(\bar{x}-x)$ scaled by their alignments $g_i$ and smoothened by the function $U(|\bar{x}-x|)$ defined in Fig. \ref{Ufunction}. This method efficiently samples the error signal along the most relevant directions determined by the look-up table boundaries. 

Observe that matrix multiplication is not required for the forward pass; it is, however,  required for the backward pass to calculate the alignment $g_i$ for each look-up table. A method of finding $g_i$ using a LUT of the form (\ref{y=Sx}) would eliminate MatMul operations and the need for GPUs during training.

In this general formulation, we have freedom of choosing the hash function $H(x)$. Useful functions are those requiring few computations, producing large partitions, allowing simple look-up table representations, and readily rendering $\bar{x}$ for any $x$. Our choice (\ref{j=H}) satisfies all these requirements. Other choices are discussed in Sect. \ref{Appendix: Polychrony}, and it seems that any choice inspired by  spike-timing patterns leads to a useful $H(x)$.

\section{Conclusion}

While both ANNs and SNNs rely on vectors, the interpretation of those vectors is what truly dictates how we manipulate them. For ANNs, the pairwise relationships between vectors—such as orthogonality and correlation—are paramount. For SNNs, however, the relationships among elements within each vector are paramount, as these relationships correspond to the different timing of spikes. This difference in interpretation leads to fundamentally different computational approaches. However, SNNs retain the most important operational aspect of ANNs: learning by gradient descent. This retention allows us to successfully reproduce deep networks, RNNs, and transformers, opening the possibility to integrate other architectural innovations developed for ANNs into SNNs.

\vspace{0.5cm}

{\bf Acknowledgement:} Oleg Sinavski and Mikhail Burtsev read the first draft of this manifesto and made me rewrite it completely. Anatoli Starostin,  Frank C. Hoppensteadt, Alfredo Maria Solano, Takeshi Kojima, Kan Hatakeyama, Assaf Touboul, Ilya Rybak, Bill MacCartney, Thomas Stahura, and Todd Hylton have made a number of useful suggestions. Special thanks to Taylor Kergan from UCSC, who provided simulation results of the ANN transformer.

\onecolumn 
\newpage 
\twocolumn

\section{Appendix (extended discussion)}

\subsection{Feature encoding capacity of ANNs}
\label{Memory representations}

Let us use the scaled dot-product attention mechanism of the standard transformer architecture to illustrate a point about capacity of ANNs and SNNs to store and manipulate vectors. The attention is based on the linear algebra observation that if we have approximately orthogonal $n$-dimensional "key" vectors $k_i$ of norm $1$ and associated with them arbitrary "value" vectors $v_i$, then the matrix
\[
M = \sum_{i=1}^N k^{\top}_i v_i
\]
stores such pairs. If a "query" vector $q$ matches some of the keys, e.g.,  $q \approx k_i$, then $qM = \sum  q k^\top_i v_i \approx v_i$, i.e., it retrieves the corresponding value vector; indeed, the dot product $q k^\top_i \approx 1$, but $q k^\top_j  \approx 0$ for other $j \neq i$. The single-head attention  
\begin{equation}
\label{softmaxAtt}
\mbox{softmax}(\frac{\ q \, k^\top}{\sqrt{n}}) v
\end{equation}
applies softmax operation to the dot products to scale up some weights of the value vectors $v_i$, making $q$ "attend" to such $k_i$. 

We can only squeeze $n$ orthogonal vectors into $\R^n$. However, for large $n$ we can squeeze more vectors if they are "nearly" orthogonal, e.g., their cosine angle is less than $\varepsilon = 1/\sqrt{n}$. The linear capacity to store such "nearly orthogonal" vectors scales like $e^{c(\varepsilon) n}$ with some function $c(\varepsilon) \approx \e^2$ (Johnson-Lindenstrauss Lemma \cite{JohnsonLindenstrauss}),  but even if it were $e^n$ it would still be infinitesimally small compared with $n!$.  

Retrieving value vectors through queries has two useful properties:
\[
\begin{array}{rl}
\mbox{robustsness: } & (k_i + \mbox{noise})M \approx v_i  \\
\mbox{superposition: } & (k_i + k_j)M \approx v_i + v_j 
\end{array}
\]
Anthropic team has a number of papers discussing these properties \cite{Toy Model of Superposition}, \cite{toy-double-descent}.

Polychronous patterns also have these properties, but express them in a different way: 
\begin{itemize}
\item {\sc Robustness}: The way we define polychronous patterns in this manifesto makes them robust to noise that preserves relative order of spikes of anchor neurons $x_a$ and $x_b$. What the noise does to the other neurons does not influence any given polychronous pattern. 
\item {\sc Superposition}: Any two polychronous patterns that involve different sets of anchor neurons trivially superpose. Overlapping patterns also superpose if they involve the same order of spikes of overlapping anchor neurons. In fact,  simple combinatorics shows that a polychronous pattern involving $n_c$ neurons in $\R^n$ will co-exist with $n!/n_c!$ other patterns. From a practical standpoint, this number is effectively infinite even when $\R^n$ is so small that the number of stored ANN patterns is just $n$. 
\end{itemize}
To illustrate superposition, let us use a popular analogy from word embeddings: Assume anchor neurons $x_1$ and $x_2$  are part of many polychronous patterns. The relationship $x_1 > x_2$ might correspond to masculine nouns such as "king", "boy", "son", etc., while the reverse order is part of "queen", "girl",  "daughter", etc.  In ANNs, we add vectors representing features to create new vectors representing new features. In SNNs, we permute order of spikes to create new feature representations. This is why ANNs rely on matrix operations, but SNNs rely on detecting timings of spikes via look-up tables.

\begin{figure}[t]
    \centering
    \includegraphics[width=0.48\textwidth]{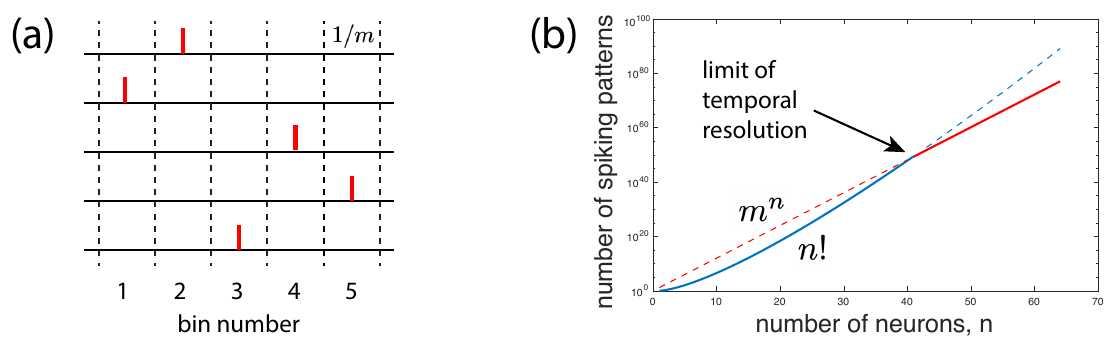}
    \caption{Calculating capacity of SNNs via quantizing the temporal interval: (a) Spikes occurring within the same bin of size $1/m$ are considered to be simultaneous. (b) The number of permutations of $n$ spikes with $m=16$ temporal bins is first dominated by $n!$ and then by $m^n$. }
    \label{quantized_spikes.pdf}
\end{figure}

\subsection{What is the capacity of SNNs?}
\label{capacity of SNNs}

Let us quantify the number of distinct spiking patterns a network of $n$ neurons, each firing one spike during a unit time interval, can exhibit. To formalize the notion of "distinct," we introduce a minimal, distinguishable time interval $\varepsilon$. Spikes whose relative latency is less than $\varepsilon$ are considered simultaneous. To simplify analysis, we quantize time into $m \approx 1/\varepsilon$ discrete bins, illustrated in Fig. \ref{quantized_spikes.pdf}a. Each spike can have $m$ positions, and a network of $n$ neurons has $m^n$ distinct spiking patterns. This number is vastly greater than the linear capacity utilized by conventional ANNs discussed in the previous section. For example, 100 neurons each firing 1 spike during 100 ms interval with $\varepsilon =1$ ms resolution, which is the temporal resolution of a typical spike-timing dependent plasticity (STDP),  could exhibit $100^{100}$ distinct spiking patterns. 

Interestingly, considering only the relative order of spikes in small networks underestimates the available number of polychronous patterns because $n! < m^n$ when $n < m$. However, when $n$ is too large, too many spikes are compressed into $m$ bins, and many of the $n!$ patterns are no longer distinct. Thus, the $m^n$ capacity serves as the outer bound for large networks; see Fig. \ref{quantized_spikes.pdf}b.

We do not want to detect all these patterns, as the required storage would exceed the capacity of the visible universe. However, we leverage this practically infinite representation capacity to create a model that never runs out of available features, even when the number of neurons $n$ is small. This approach offers major savings on compute and memory bandwidth. Our choice of the hashing function $j=H(x)$ (Eq. \ref{j=H}) fulfills this goal: we can detect exactly $2^{n_tn_c}$ distinct polychronous patterns, and we can make this number as large as we want by increasing $n_c$ without  affecting the active memory bandwidth or computational demand.

To summarize, the linear encoding capacity of ANNs is $e^{c(\e)n}$ and that of SNNs is $m^n$ for large $n$. Since $e^{c(\e)} \approx 1$ whereas $m \approx 1/\e$, the encoding capacity of SNNs dwarfs that of ANNs.

\subsection{Polychrony} 
\label{Appendix: Polychrony}

The idea of looking at the order of spikes rather than their precise timing was explored by Simon Thorpe \cite{Rate vs Temporal,SimonThorpe} almost 30 years ago to explain fast visual processing. He called it "rank order coding." In his view, the first few spikes, or even the very first spike, convey all the information about the input. In contrast, we have to account for the relative order of all spikes to detect the correct polychronous pattern and determine which index to use in the "polychronous" look-up tables. Flipping the order of the last two spikes completely changes the identity of the polychronous pattern, resulting in a different index. 

We were quick to equate a polychronous pattern with a permutation of spikes. While related, these two approaches are distinct: The same order of spikes, but with different timings, might correspond to a different polychronous pattern in a biological network; however, it would be classified as the same pattern in this manifesto. Conversely, if a polychronous pattern has two neurons firing approximately at the same time with some small noisy jitter, this jitter might reverse the order of the two spikes and result in the detection of two distinct patterns, even though they are indistinguishable from the biological point of view.

Figure \ref{quantized_spikes.pdf} offers a mechanism to detect polychronous patterns based on (quantized) spike-timing and not relative latencies: Let the function bin$(x_k)$ give the bin number of the spike with latency $x_k$, i.e., the number between $1$ and $m=5$ in the figure. Then, we concatenate bin numbers of anchor neurons $a_i = (a_{i1}, \ldots, a_{in_c})$ in base $m$ to get the index $j$, i.e.,  
\[
j = H_i(x) = \sum_{r=1}^{n_c} m^{r-1} \{\mbox{bin}(x_{a_{ir}})-1\}\;.
\]
Instead of $2^{n_tn_c}$, this hashing function detects $m^{n_tn_c}$ distinct polychronous patterns. The learning rule through gradient descent retains the same form, except min$(|u_{ir}|)$ now corresponds to $x_{a_{ir}}$ that is closest to the edge of its bin. The toy example in the next section corresponds to $m=2$ bins.

\begin{figure}[t]
    \centering
    \includegraphics[width=0.45\textwidth]{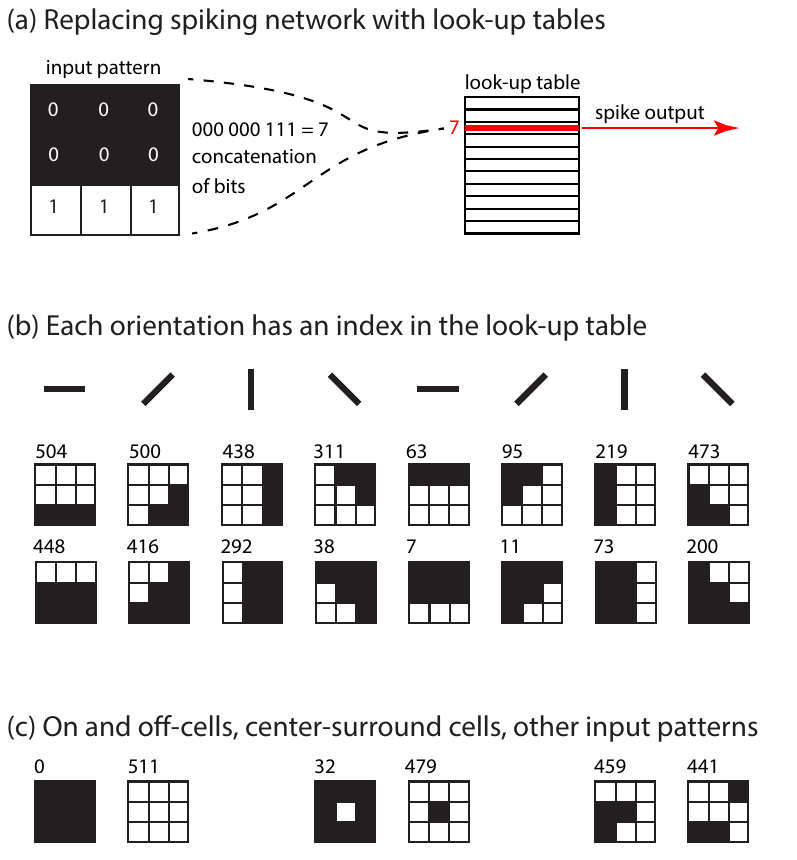}
    \caption{The equivalence between spiking networks and look-up tables: (a) A binary input is treated as bits to form an index into a look-up table; here binary 111 is decimal 7. From the outside observer, the input results in neuron "7" to fire a spike. (b) All orientations have binary representations corresponding to specific entries in the table. (c) The look-up table represents all possible input features, including some that might have no meaning, such as 459 and 441.}
    \label{spiking}
\end{figure}

\subsection{Spiking networks as look-up tables}

A major pillar of this manifesto is the claim that spiking networks are nature's way of implementing look-up tables. To illustrate this, consider a toy example of building edge detectors tuned to various orientations of 3x3 patches, as shown in Fig. \ref{spiking}. This is similar to what filters in the first layer of convolutional neural networks do. A traditional approach, inspired by the orientation-selective cells of the mammalian visual cortex, is to have neurons tuned to various orientations through synaptic strengths. Sufficiently strong lateral inhibition ensures that only the neurons receiving the strongest input transmit to the next processing layer, while most remain silent. If one neuron, out of 100, fires a spike, only the synapses from that neuron are used in the next processing layer. Synapses from all other neurons are not used at all, at least not until those neurons fire. From an outside observer's point of view, the input magically selects a vector of values out of 100 possible vectors. This is precisely what an index into a look-up table does.
 
Figure \ref{spiking} illustrates another approach to process visual input using look-up tables: We treat each input patch as a binary vector and use it as an index into a look-up table, which contains synaptic values to the next layer of processing. For example, the horizontal edge in the figure results in the binary 000\,000\,111, which is decimal 7, causing neuron 7 to fire a spike. Other orientations produce other indices, with a total of $2^9 = 512$ neurons tuned to all possible inputs, including the orientations in Fig. \ref{spiking}b and other features shown in Fig. \ref{spiking}c.  For any given input, this mechanism selects precisely which output neuron will fire a spike. The input does not have to be binary or small. For a general input $x\in\R^n$, we select $n_c$ anchor neurons $x_{a_1}, \ \ldots, \ x_{a_{n_c}}$, and then use the hash function 
\[
j=H(x) = \mbox{concat}(x_{a_1}>0, \ \ldots, \ x_{a_{n_c}}>0)\;.
\]
Incidentally, this is what we do to create indices for the positional encoders $PE\in\R^p$ in the SNN attention mechanism when $p$ is small. 

This toy mechanism illustrates all the important aspects of SNNs presented in this manifesto: (a) extreme processing efficiency -- that is, getting the answer without simulating individual neurons, (b) built-in sparsity, and (c) the detection of all possible orientations and features at no additional cost, even those with no obvious meaning that might never be encountered in the natural input.  

It is astonishing that the interpretation of spiking networks as look-up tables -- where firing a spike is equivalent to retrieving an entry -- has never been proposed before. Even more astonishing is the thought that this interpretation might have been proposed some time ago but was ignored by the AI community.

\begin{figure}[t]
    \centering
    \includegraphics[width=0.46\textwidth]{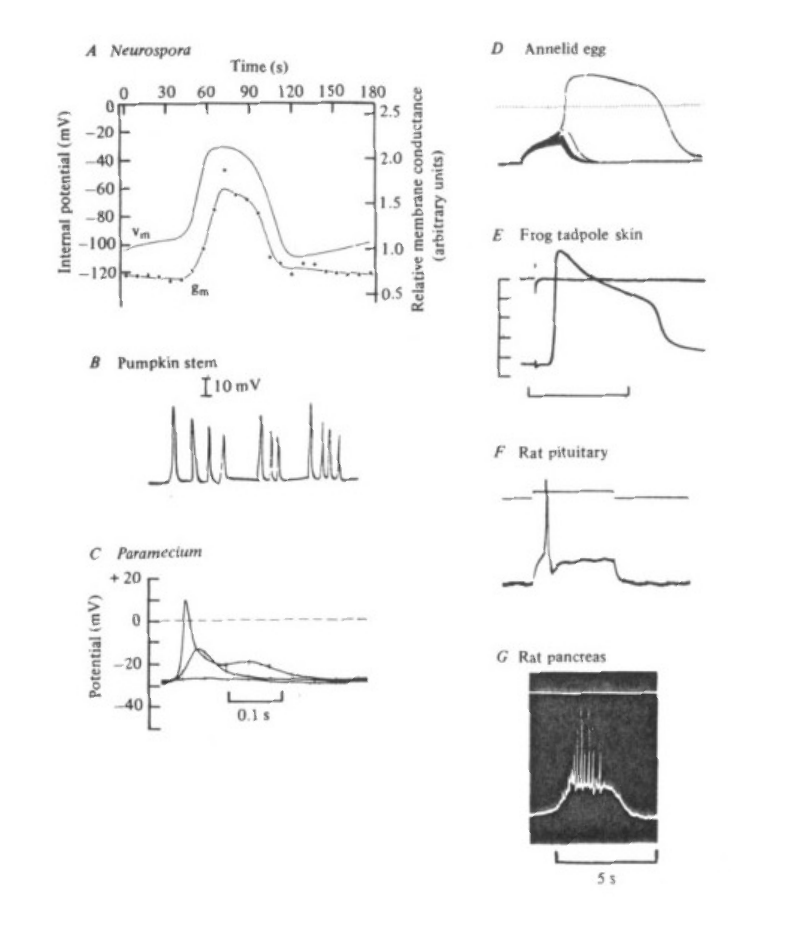}
    \caption{Examples of spikes in non-neural cells (from \cite{Neurobiology}).}
    \label{Shepherd}
\end{figure}

Nature did not invent spiking neurons just for information processing. Spiking is ubiquitous in non-neural cells, and even pumpkins do that; see Figure \ref{Shepherd}B. When nature needed to implement look-up tables for efficient encoding and processing, it simply used the tools at its disposal.

\subsection{Latency} 
\label{Appendix: Latency}

We were very cavalier to replace a spiking pattern with a vector of latencies $x\in\R^n$. An implicit assumption was that each neuron fires exactly one spike during a certain interval. What if a neuron is quiet? Most are. In this case, we can treat such a neuron as having a spike with some large latency. What if a neuron fires a doublet or a burst? In that case, we just take the first spike in the burst. 

\begin{figure}[t]
    \centering
    \includegraphics[width=0.25\textwidth]{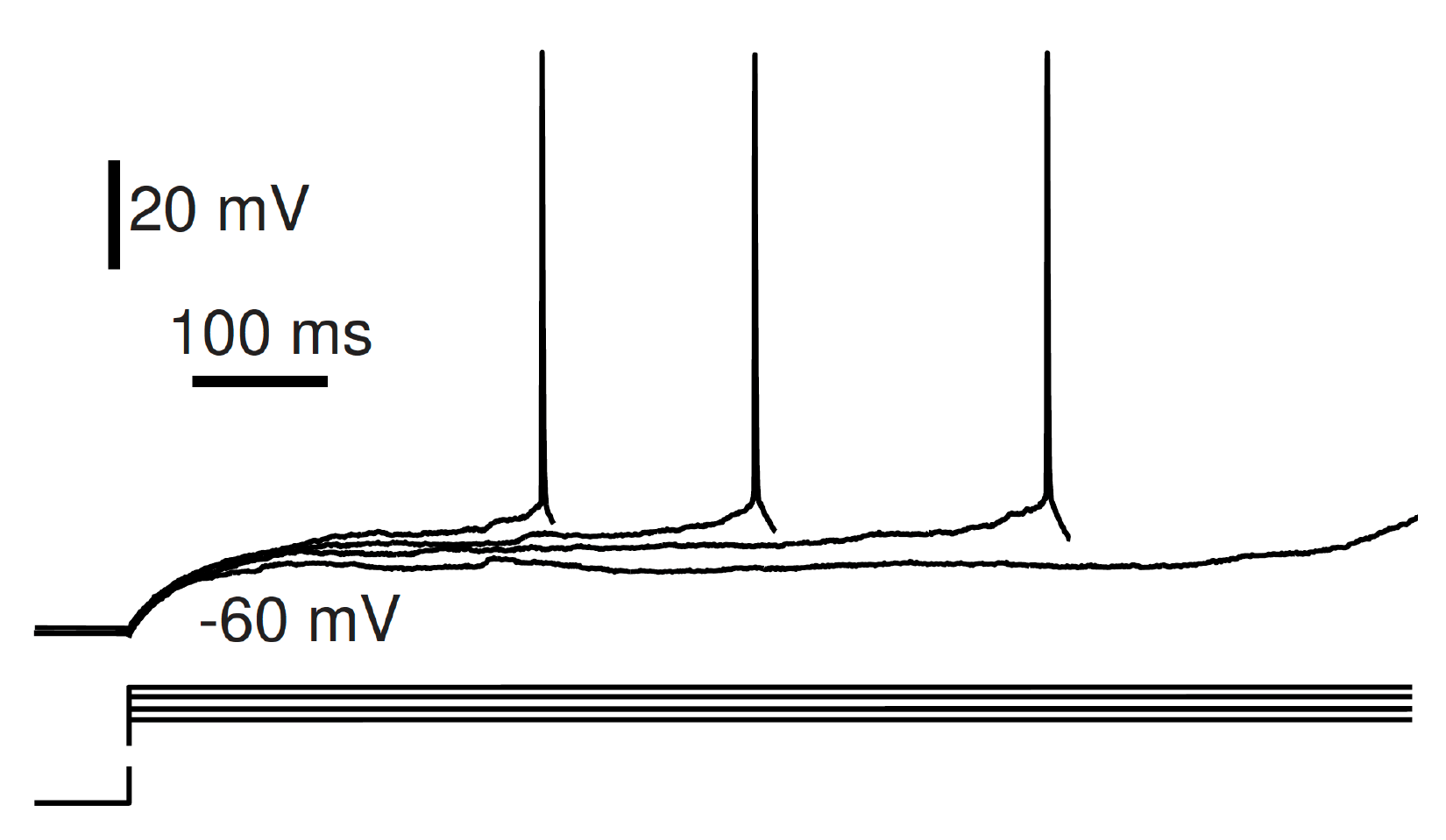}
    \caption{In vitro recordings of the pyramidal neuron of layer 2/3 of a rat’s visual cortex show that it encodes the strength of the input into the latency: shorter latencies for stronger injected currents \cite{dsn}.}
    \label{input2latency}
\end{figure}

We also treated the synaptic output from look-up tables as latencies, and used simple addition to get the latency of the postsynaptic neurons. With this respect, the synaptic values have units of time. To follow the electrophysiology of spiking neurons, we would need to add synaptic values to the postsynaptic neuron's input current or synaptic conductance, and then calculate the time to the first spike. However,  cortical pyramidal cells have Class 1 excitability \cite{dsn} and encode the strength of the input into the latency of their response in a monotonic manner. The stronger the input is, the earlier they fire, as shown in Fig. \ref{input2latency}. If we are only interested in the relative order of spikes, then treating the postsynaptic values, $y_k$, as inputs or as latencies would not change the relative order; it would only change how we depict spikes in our figures, with larger $y_k$ firing first or the mirror image with larger $y_k$ firing last. 

Let's assume there is another, nonlinear, way to accumulate inputs from different look-up tables into a postsynaptic neuron:
\[
y = f(S_{1x}, \ldots, S_{n_tx})\;.
\]
If all synaptic vectors $S_{ix}$ are near zero, we can linearize $f$ near $S_{ix}=0$ and end up with a constant term $f(0,\ldots,0)$ plus linear summation of vectors $S_{ix}$ multiplied by a constant Jacobian matrix, which could be absorbed into the synaptic values $S_{ix}$. For the nonlinear function above to be truly distinct from linear summation, it must have some symmetry or other constraint resulting in zero Jacobian matrix, e.g., $f_k = \sum_i (s_{ixk})^2$.

\begin{figure}[t]
    \centering
    \includegraphics[width=0.36\textwidth]{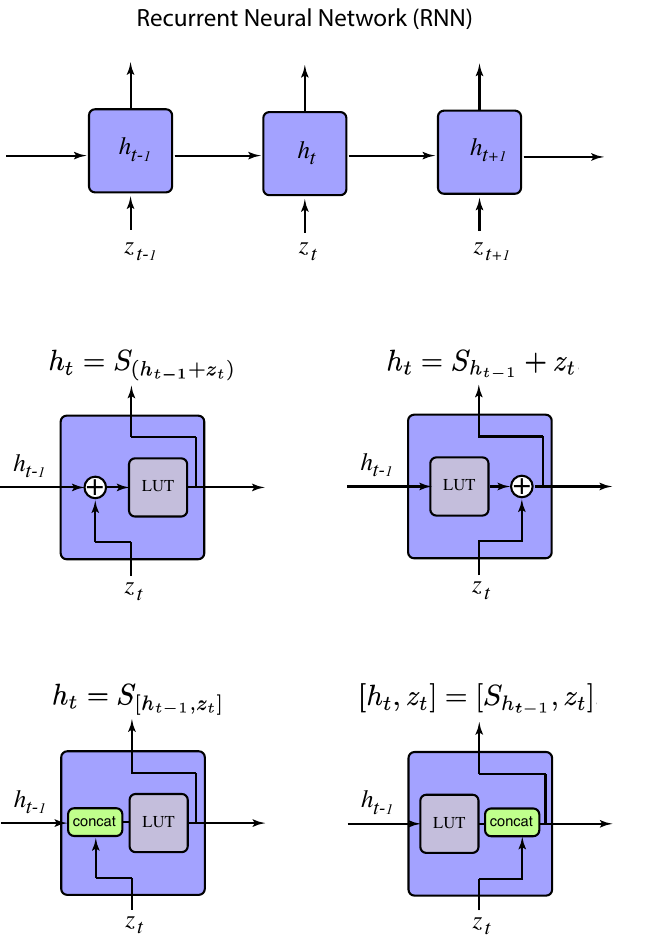}
    \caption{Implementing a typical RNN (top) as an SNN provides four choices of how to combine the input vector and the hidden state.  }
    \label{RNN.pdf}
\end{figure}

\subsection{Spiking RNN}
\label{Section: Spiking RNNs}

The conversion of an RNN into a spiking RNN would be straightforward, were it not for the four distinct ways the input could be combined with the hidden state to replace matrix multiplication, as we summarize in Fig. \ref{RNN.pdf}. Combining vectors via concatenation is a unique SNN feature; indeed, applying matrix multiplication to concatenated vectors in ANNs is the same as multiplying the vectors by their own matrices and then adding up the results, producing no new functionality. Concatenation of the input vectors is important for the SNN implementation of attention in transformers.

\begin{figure}[t]
    \centering
    \includegraphics[width=0.4\textwidth]{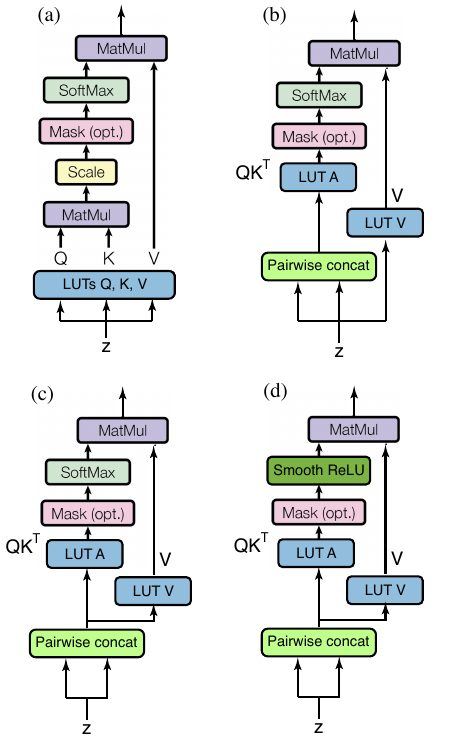}
    \caption{Gradual removing of matrix multiplication and softmax operation in attention module. }
    \label{other_attention.pdf}
\end{figure}

\subsection{Other implementations of attention}
\label{Appendix: Attention}

Let us list various implementations of the transformer attention module by gradually removing matrix multiplication and softmax operations. We drop positional encoders for simplicity. 

\begin{itemize}
\item
The most straightforward implementation is to have LUTs of the form (\ref{y=Sx}) for each query, key, and value vector
\[
Q_i = W^Q_{z_i}, \ \ \ K_i = W^K_{z_i}, \ \ \ V_i = W^V_{z_i}\;, 
\] 
and then use dot-product with softmax (\ref{attention}) to calculate the output of the attention module. That is, we only replace the first MatMul box with LUTs in Fig. \ref{other_attention.pdf}a, leaving the rest of the attention architecture unchanged. This version has matrix multiplication and softmax bottleneck, and hence is undesirable. 

\item Going further, we do not need to compute $n^2_{inp}$ dot products $Q_iK^\top_j$ for all pairs of queries and keys. We do not even need to compute $Q_i$ and $K_j$ vectors. Instead, we make pairwise concatenation $[z_i, z_j]$ of the embeddings and feed them into a learnable LUT $A_z$ of the form (\ref{y=Sx}) to get raw attention scores:
\[
a_{ij} = A_{[z_i, z_j]} \ \ \ \ \ \mbox{($=QK^\top$)}
\]
Then, we pass these through the softmax operation and apply them to the pre-computed value vectors $V_i$, as in Fig. \ref{other_attention.pdf}b. Interestingly, the performance of the model is better when the value vectors are also computed through a pairwise concatenated LUT 
\[
V_{ij} = W^V_{[z_i, z_j]}\;,
\]
as in Fig. \ref{other_attention.pdf}c. Using the V-index caching, one can make this linear in $n_{inp}$.

\begin{figure}[t]
    \centering
    \includegraphics[width=0.25\textwidth]{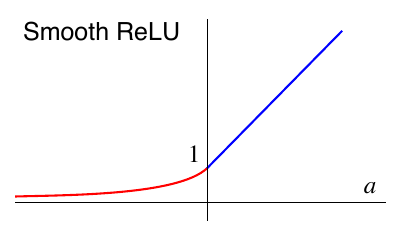}
    \caption{Smooth positive Rectifying Linear Unit $R(a)=1/(1-a/T)$ for $a\leq 0$ and $R(a) = 1+a/T$ for $a>0$, where $T>0$ is the temperature as in softmax. }
    \label{smoothReLU.pdf}
\end{figure}

\item
We can drop the softmax operation and instead pass the raw attention scores $a_{ij}$ through a smooth ReLU function of the form in Fig. \ref{smoothReLU.pdf}, then multiply by the value vectors, so that the output of the attention module is 
\[
x_i = z_i + \sum_{j=1}^{i-1} R(a_{ij}) V_{ij}\;,
\] 
as in Fig. \ref{other_attention.pdf}d. 
The long positive tail of $R(a)$ for $a<0$ is important here: it induces competitive learning dynamics by pushing most of the raw attention scores $a_{ij}$ into the negative domain, especially for temperature $T<1$. The multiplication by $R(a_{ij})$ can be removed during inference if we quantize $R(a_{ij})$ and incorporate it into the look-up table calculation of $V_{ij}$. 

\item
Even simpler implementation of attention is to replace function $R$ above with learnable non-negative positional weights $a_{i-j}$, i.e., 
\[
x_i = z_i + \sum_{j=1}^{i-1} a_{i-j} V_{ij}\;.
\]
The weights "help" to decrease contributions of value vectors at distant relative positions. 
\end{itemize}

The implementation via smooth ReLU $R(a)$ above actually performs better than the completely linear implementation (\ref{x=Vz}) used in the main part of the manifesto. This seems counterintuitive because both $a_{ij}$ and $V_{ij}$ depend on the same pair of variables, $z_i$ and $z_j$, so a combined look-up table $V_{[z_i, z_j]}$ has access to exactly the same information. For some reason, the network performs better when it learns the magnitudes (weights $a_{ij}$) of the value vectors separately from their direction $V_{ij}$.

\subsection{Hybrid approach: LUT for vanilla Transformer}
\label{hybrid approach}

In modern LLMs, each embedding is represented by a 16,384-dimensional vector. It is an overkill to have many tables with a lot of comparisons to sample from all 16,384 neurons, especially because spiking patterns need fewer neurons to have large encoding capacity. Nevertheless, we can still merge the SNN framework and the vanilla transformer to create a hybrid architecture:  We can use anchor vectors $c_{ir} \in \R^{16384}$ to dot-multiply $x \in \R^{16384}$ to get positive or negative values for the look-up indices. That is, we can use $u_{ir} = c_{ir} \cdot x$ in the function (\ref{j=H}), a method inspired by hyperplane locality-sensitive hashing \cite{HLSH}. 

As a result, each look-up table splits $\R^{16384}$ into $2^{n_c}$ cones. If all anchor vectors are linearly independent, the entire embedding space is partitioned into $2^{n_c n_t}$ cones, which easily exceeds ANN's linear coding capacity described in Sect. \ref{Memory representations}. The anchor vectors themselves can be trained through error backpropagation
\[
\frac{\partial \L}{\partial c_{i}} =  U'( c_{i} \cdot x) \, x \, g_{i}\;,
\]
where $c_i$ is the anchor vector that results in the smallest $|c_{ir} \cdot x|$ and $g_{i}$ is defined by (\ref{gi}). Not surprisingly, performance of this learning rule is better than that of the standard rule (\ref{dLdx}). It simply has more trainable parameters. We do not use it in our simulations because it requires matrix multiplications.      

The models we describe in this manifesto are a special case when $c_{ir}$ has all zeros, one +1 (in location $a_{ir}$ ), and one -1 (in location $b_{ir}$). Therefore, we can insert SNNs into existing ANNs. In particular, this approach would allow the use of LUTs and V-index caching to get better scaling of the attention module in the standard dot-product ANN transformer. This might be a beachhead to introduce spiking networks to the wider ML community.

\subsection{Synaptic plasticity vs. structural plasticity}
The current learning mechanism in the SNN defined by Equation (\ref{y=Sx}) modifies only the synaptic strengths $s_{ijk}$. The anchor neurons that define the polychronous patterns, $x_a$ and $x_b$, are randomly chosen at initialization and fixed. What if we can modify these neurons, essentially rewiring the random connections in the spiking network, thus creating structural plasticity.  

One guiding principle for modifying anchor neurons might be to ensure that all rows of look-up tables (all $j$) have equal probability of being used. For example, if two comparisons always result in the same sign, half of the look-up table rows are never used. Another principle focuses on mutual information: each anchor pair $u_{ir} = x_{a_{ir}} - x_{b_{ir}}$ partitions the look-up table into two halves corresponding to $u_{ir}>0$ and $u_{ir}<0$. If these halves have similar synaptic values, this pair brings no new information to the rest of the anchor neurons and should be replaced. Yet another method could involve the hyperplane locality-sensitive hashing described in Sect. \ref{hybrid approach}. If the loss function ensures that the anchor vector $c\in\R^n$ is sparse (with only two non-zero entries), one can pre-run the hybrid model first to find the most informative pairs of anchor neurons and use them to initialize the non-hybrid SNN model.

\subsection{Parameter-efficient fine-tuning}
\label{SNN fine-tuning}

Fine-tuning is the process of specializing a large, pre-trained model on a new, smaller dataset (e.g., medical text or legal documents) to optimize its performance for a specific task.  To avoid high computational cost, techniques like LoRA (Low-Rank Adaptation, \cite{LoRA}) freeze the original ANN model weights and only train small, dedicated "adapter" parameters.

The architecture presented in this manifesto offers two distinct ways to achieve this kind of parameter-efficient fine-tuning on new data without modifying the existing, vast set of learned synaptic values. We denote these methods as $n_t+1$ and $n_c+1$. Both methods increase the model's memory footprint by a marginal amount, $2^{n_c}$ rows, but allow the model to learn new patterns that gradually shift the network's output without catastrophic forgetting.

\begin{itemize}
\item \textsc{New Table ($n_t+1$):} This approach involves adding a single new look-up table to the existing set of $n_t$ tables, initialized with zero synaptic values, $S_{n_t+1}\equiv 0$. The model then passes gradients backward only through this single new table. Since the table starts at zero, it introduces no change to the current forward pass, but it allows the network to learn a separate, dedicated set of synaptic values optimized solely for the new fine-tuning task.

\item \textsc{New Comparison ($n_c+1$):} This method modifies an existing table by adding one new anchor-pair comparison, $x_a > x_b$, which effectively doubles the number of rows from $2^{n_c}$ to $2^{n_c+1}$. Crucially, the $2^{n_c}$ new synaptic rows are initialized with a copy of the old $2^{n_c}$ rows. This means the initial output remains identical. The gradients are then passed backward only through this modified table, allowing the model to refine and split the existing $2^{n_c}$ patterns into $2^{n_c+1}$ subtler patterns tailored to the new data.
\end{itemize}

\subsection{More learning rules}
\label{Backprop without MatMul}

We can make the surrogate equation (\ref{surrogate}) more complicated and include all pairs $u_{ir} = x_{a_{ir}} - x_{b_{ir}}$,  
\[
y =  \frac{1}{n_c} \sum_{i,r} \left( S_{ix}  +  U(u_{ir}) ( S_{i\bar{x}} - S_{ix})  \right) \;. 
\]
The equation results in the gradient vectors 
\begin{equation}
\label{all_a}
\frac{\partial \L \ }{\partial x^l_{a^l_{ir}}} = - \frac{\partial \L \ }{\partial x^l_{b^l_{ir}}} = U'(x^l_{a^l_{ir}} - x^l_{b^l_{ir}}) \frac{1}{n_c}  \frac{\partial \L \ \ }{\partial x^{l+1}} \cdot \left(S^l_{i\bar{x}} - S^l_{i{x}}\right) 
\end{equation}
that involve more computational resources, but it does not result in better learning performance.  This is equivalent to considering many perturbation points $\bar{x}$ in Fig. \ref{generalization} that map to different hash buckets. 

Alternatively, we can simplify the surrogate equation (\ref{surrogate}), e.g., by ignoring the flipped synaptic weights $S_{i\bar{x}}$. The surrogate function, 
\[
y = \sum_{i=1}^{n_t}  S_{ix} (1-U(x_{a_{i}} - x_{b_{i}}))\;,
\]
just scales down the vector of synaptic values $S_{ix}$ by the smallest difference $u_i = x_{a_{i}} - x_{b_{i}}$. The simplest interpretation is that the smaller the difference $|u_i|$ is, the more unsure we are about the synaptic values, so it is better to scale them down so that they do not interfere with synaptic values from the other look-up tables. This surrogate function also results in a viable learning rule
\begin{equation}
\label{no x bar}
\frac{\partial \L \ }{\partial x^l_{a^l_{i}}} = - \frac{\partial \L \ }{\partial x^l_{b^l_{i}}} = - U'(x^l_{a^l_{i}} - x^l_{b^l_{i}}) \;  \frac{\partial \L \ \ }{\partial x^{l+1}} \cdot S^l_{ix}\;.  
\end{equation}
This rule is equivalent to using $0$ instead of $S_{\bar{j}}$ in Fig. \ref{generalization}. Because it has smaller memory bandwidth requirement (no need to load $S_{i\bar{x}}$) but almost the same performance as our standard rule, it might be appropriate in the computationally intensive cases like the attention block in SNN Transformers. 

We can take this to the extreme and stipulate that we are only going to consider a single look-up table at layer $l$ that produces the smallest latency difference $|u^l_{ir}|$ in the entire layer. That is, it is the only one that could flip its sign under the minimal perturbations of $x^l$.  The gradient equation has the same form as (\ref{dLdx}), 
\begin{equation}
\label{the minimal learning rule}
\frac{\partial \L \ }{\partial x^l_a} = - \frac{\partial \L \ }{\partial x^l_b} = U'(x^l_a - x^l_b) \;  \frac{\partial \L \ \ }{\partial x^{l+1}} \cdot (S^l_{i\bar{x}} - S^l_{i{x}})  
\end{equation}
where $i$ denotes the table and $a$ and $b$ denote that specific pair, $u^l_{ir} = a^l_{ir} - b^l_{ir}$, that has the minimal latency difference $|u^l_{ir}|$ in the entire layer.  Among all learning rules described above, this rule has the best computational performance (speed), but the worst learning performance. 

\begin{figure}[t]
    \centering
    \includegraphics[width=0.45\textwidth]{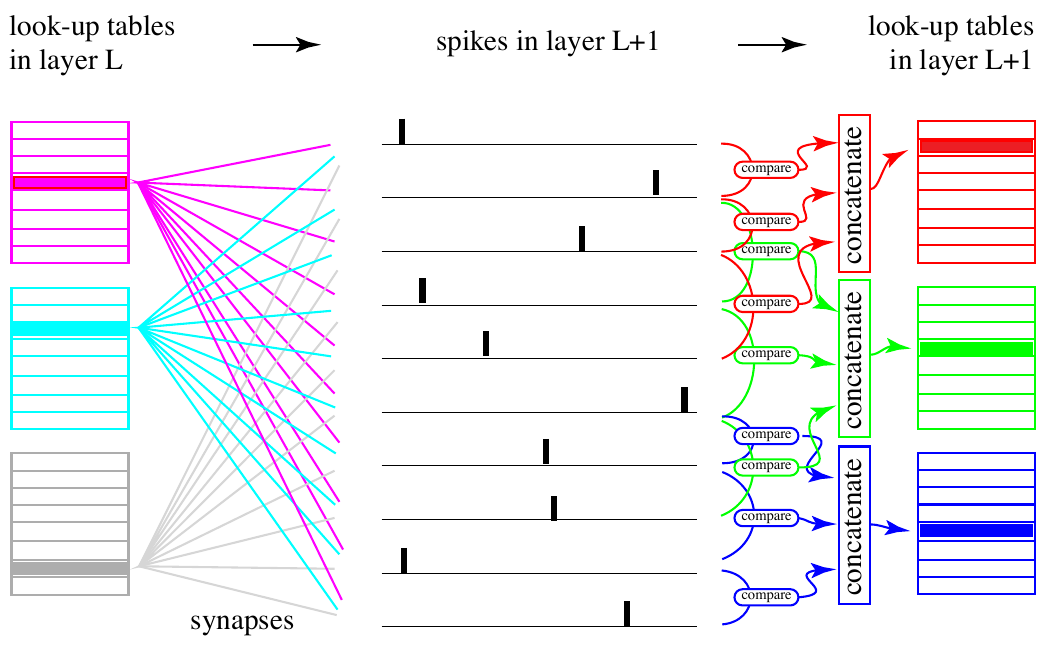}
    \caption{A view of a spiking network as a system where polychronous patterns in one layer are transformed into polychronous patterns in the next layer.}
    \label{poly2poly}
\end{figure}

{\em Spiking backprop}: Let us combine the last two rules, i.e., consider only $S_{ix}$ and ignore $S_{i\bar{x}}$ and consider only one minimal pair (in table $i$) per entire layer, to illustrate an important point. If we denote the gradient as 
\[
h^l = \frac{\partial \L}{\partial x^l_a}\;.
\]
then the combined rule can be written as 
\[
h^l = U'(x^l_{a^l} - x^l_{b^l}) \; (s^l_{i{x}b^{l+1}} - s^l_{i{x}a^{l+1}}) \; h^{l+1}\;. 
\]
Thus, a single number is transmitted from layer $l+1$ to the layer $l$ to convey the error gradient. Moreover, the error gradients propagate back not from neurons in layer $l+1$ to neurons in layer $l$, but from one look-up table in layer $l+1$ to all look-up tables in layer $l$ without involving the neurons, as in Fig. \ref{poly2poly}. The original learning rule can also be rewritten in this form, requiring propagation of $h^{l+1}_i$ from each look-up table in layer $l+1$ to each look-up table in layer $l$. 

It is also interesting because $h^{l+1}$ acts as a global signal to the entire layer $l$ to govern synaptic plasticity there, suggesting a spiking solution to the biological plausibility of error backpropagation in the brain  \cite{backpropinthebrain}. Additionally, going from $h^{l+1}$ to $h^l$ does not involve matrix multiplication, suggesting that the same spiking mechanism can be used for both inference and learning. 

Finally, notice that the computational resources required for learning -- specifically, to determine the minimal $u^l_i$ in each layer and then to update two outgoing synaptic values, $s^l_{ixa^{l+1}}$ and $s^l_{ixb^{l+1}}$, for each table $i$ --  are negligible compared with the vector additions (\ref{y=Sx}) required for inference.

\subsection{Relationship to Mixture-of-Experts}
\label{Section: MoE}

SNNs can be considered as the simplest examples of Mixture-of-Experts (MoE) models \cite{Shazeer} or sparse expert models \cite{Sparse MoE}. In all these architectures, a router selects which subset of the model acts as an "expert" for any particular input $x$. In our framework, the router is the function (\ref{j=H}) that selects which row, $j$, of each look-up table to use for processing. Since the number of selected rows and their dimension are always the same, our routing has perfect load balance.

The core benefit shared with MoE is that the total number of stored parameters far exceeds the number of parameters used for any single computation. While the SNN has a synaptic footprint proportional to the total size of all look-up tables $n_t2^{n_c}n$, the active processing is negligible. At any given time, the SNN router only retrieves $n_t$ rows (one per table), resulting in a minimal, constant computational load, similar to MoE  that activate only a few expert blocks. This architecture allows the model to scale its memory and representation capacity exponentially, $2^{n_tn_c}$, while maintaining low, linear computational complexity and minimal memory bandwidth demands, which is why MoE and our SNN are significantly more efficient than dense ANN models.

\subsection{Relationship to finite-state machines}
\label{Section: finite-state machines}
 
In this manifesto, we treat look-up tables as transformations from input latency vectors $x$ to output latency vectors $y$. However, a complementary view, depicted in Fig. \ref{poly2poly}, is to treat the latency vectors as local auxiliary variables (not shown) and to consider indices of look-up tables as the states of the network. The indices in layer $L$ determine the indices in layer $L+1$. From this perspective, we have a network generating spiking patterns but without spiking neurons.  
 
Figure \ref{poly2poly} suggests that we can view SNN dynamics as a finite-state machine, where each state is described by the $n_t$ indices of the look-up tables. The dimension of this machine is finite, but quite big, i.e., $2^{n_tn_c}$ per layer, e.g., $2^{64\times 10} \approx 10^{192}$ in the spiking RNN in Sect. \ref{Experiments}. Since $10^{192}$ is an infinite state space from practical point of view, the comparison with finite state machines might not be useful. 
 
However, it is tempting to create rules that would describe the transition between the set of $n_t$ indices in one layer to the indices in the next layer, i.e., from $2^{n_tn_c}$ polychronous patterns in one layer to polychronous patterns in the next layer, without modeling the latencies $x_k$ explicitly.

\subsection{Relationship to random forests and ferns}
\label{Section: forests and ferns}

Consider a single-layer SNN (\ref{y=Sx}) and denote $p_{ijk} = $softmax$(s_{ijk})$. Then, we can treat each $y_k$ in 
\[
y_k = \prod_i p_{ixk}
\]
as the probability of the $k$th class in a semi-naïve Bayesian classifier. Since each look-up table considers all possible permutations of latencies of its anchor neurons, the classifier above is called random fern \cite{RandomFern}. 

There have been attempts to build deep networks of such classifiers \cite{DeepForest}, \cite{DeepFerns}, but all these efforts acknowledge the non-differentiable nature of the models. As a result, they focus on non-gradient-descent methods. 

\subsection{Relationship to transformer quantizations}
\label{Section: Transformer quantizations}

There have been attempts to recast the transformer architecture using spikes, including Spike-driven Transformer \cite{SpikeDriven}, SpikeLLM \cite{SpikeLLM}, SpikeGPT \cite{SpikeGPT}, SpikeBert \cite{SpikeBert}, SpikeFormer \cite{SpikeFormer}, BrainTransformers \cite{BrainTransformers}, and the Loihi 2 implementation of a 370M-parameter LLM \cite{LLMonLoihi2}.

None of these attempts take advantage of the combinatorially explosive encoding capacity of SNNs. Instead, they all can be viewed as an effort to quantize data into "ternary" spikes (-1, 0, and +1) and use sparsity to improve computational efficiency. Not surprisingly, matrix multiplication of non-zero elements is replaced by a nearly equal number of additions and subtractions \cite{MatMulFree}, but the underlying ANN framework remains the same. In particular, none of them incorporate LUTs.

\begin{figure}
    \centering
    \includegraphics[width=0.4\textwidth]{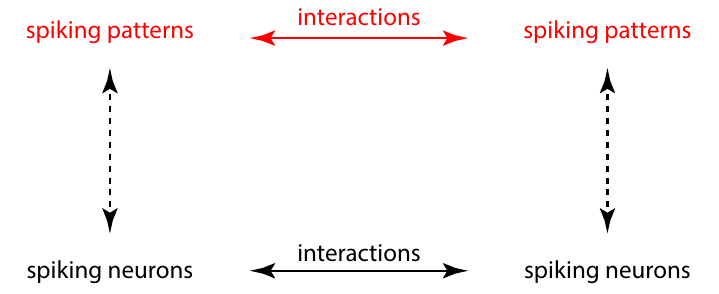}
    \caption{Black: Interactions between spiking neurons give rise to various spiking patterns. Red: Instead of modeling direct interactions between such neurons, this manifesto considers direct interactions between spiking patterns.}
    \label{CoverSpace}
\end{figure}

\subsection{Relationship to neuromorphic systems}
\label{Section: neuromorphic}

Among the many neuromorphic programs, the most notable are TrueNorth by IBM \cite{TrueNorth}, Loihi by Intel \cite{Loihi2}, and SpiNNaker by ARM creator Steve Furber at the University of Manchester \cite{Spinnaker2}. Their processors show unprecedented energy efficiency due to the sparse, event-based activity of spiking networks.

However, all these neuromorphic efforts had started before transformers were invented and even before deep learning was popularized by AlexNet \cite{AlexNet}. As a result, the processors were not designed with deep learning or transformer architectures in mind. The hope was that somebody would figure out how to use these neuromorphic architectures to take advantage of the combinatorially explosive representation capacity of spiking networks. This may indeed happen in the future, but the lesson from this manifesto is that it is faster to model "spiking patterns to spiking patterns" interactions (at the top of Fig. \ref{CoverSpace}) rather than "neuron to neuron" circuits (at the bottom of the figure). This is similar to the general trend in ANNs, where we are no longer thinking about individual neuron-to-neuron connections, but instead about "vector to vector" interactions.


\begin{thebibliography}{1}

\scriptsize

\bibitem{Hardware Lottery}
S. Hooker (2020) \emph{The Hardware Lottery}, Arxiv, 2009.06489 


\bibitem{JohnsonLindenstrauss}
W. B. Johnson and J. Lindenstrauss (1984)
{\em Extensions of Lipschitz mappings into a Hilbert space},
Contemporary Mathematics, 26, 189–206

\bibitem{CerCor}
E.M. Izhikevich, J.A. Gally, and G.M. Edelman  (2004) {\em
Spike-Timing Dynamics of Neuronal Groups,}
Cerebral Cortex, 14:933-944

\bibitem{PNAS}
E.M. Izhikevich and G.M. Edelman  (2008) {\em
Large-Scale Model of Mammalian Thalamocortical Systems,}
PNAS, 105:3593-3598

\bibitem{Vaz2020}
Alex Vaz et al. (2020)
{\em Replay of cortical spiking sequences during human
memory retrieval}
Science 367, 1131--1134

\bibitem{Vaz2023}
Alex Vaz (2023)
{\em Backbone spiking sequence as a basis for
preplay, replay, and default states in human
cortex}
Nat Communication Aug 7;14(1):4723

\bibitem{XieNature}
Weizhen Xie et a. (2024) 
{\em Neuronal sequences in population bursts encode information in human cortex}, 
Nature 635: 935--942 

\bibitem{Polychronization}
E. M. Izhikevich (2006) \emph{Polychronization: Computation with Spikes}. Neural Computation, 18:245-282

\bibitem{TrueNorth}
F. Akopian et al. (2015) \emph{TrueNorth: Design and Tool Flow of a 65 mW 1 Million Neuron Programmable Neurosynaptic Chip}, IEEE Transactions on Computer-Aided Design of Integrated Circuits and Systems, 10.1109/TCAD.2015.2474396

\bibitem{Loihi2}
https://www.intel.com/content/www/us/en/research/neuromorphic-computing-loihi-2-technology-brief.html

\bibitem{Spinnaker2}
C. Mayr at al. (2019) \emph{SpiNNaker 2: A 10 Million Core Processor System for Brain Simulation and Machine Learning}, 	arXiv:1911.02385

\bibitem{McPitts}
W.S. McCulloch and W. Pitts (1943). A logical calculus of the ideas immanent in nervous activity. The Bulletin of Mathematical Biophysics, 5(4), 115–133.

\bibitem{neuromorphic}
D. Kudithipudi et al. (2025)
{\em Neuromorphic computing at scale},
Nature, 637:801--812

\bibitem{LSH}
Piotr Indyk and Rajeev Motwani (1988) Approximate nearest neighbors: towards removing the curse of dimensionality,  {\em STOC '98: Proceedings of the thirtieth annual ACM symposium on Theory of computing}, 604--613

\bibitem{dsn}
E.M.~Izhikevich (2007) \emph{Dynamical Systems in Neuroscience: The Geometry of Excitability and Bursting}. Springer-Verlag, N.Y.

\bibitem{Neftci2019}
Emre O. Neftci, Hesham Mostafa, and Friedemann Zenke (2019)
{\em Surrogate Gradient Learning in Spiking Neural Networks},
arXiv:1901.09948

\bibitem{spikingbackprop}
J. K. Eshraghian et al. (2023) {\em Training Spiking Neural Networks Using Lessons From Deep Learning}, in Proceedings of the IEEE, vol. 111, no. 9, pp. 1016-1054

\bibitem{resnet}
Kaiming He, Xiangyu Zhang, Shaoqing Ren, and Jian Sun. (2016) {\em Deep residual learning for image recognition}. In Proceedings of the IEEE Conference on Computer Vision and Pattern Recognition, pages 770–778, 2016.

\bibitem{LSTM}
Sepp Hochreiter and Jurgen Schmidhuber (1997)
{\em Long Short-Term Memory}, 
Neural Computation, 9(8):1735--1780

\bibitem{LSTM2}
F A Gers, J Schmidhuber, and F Cummins (2000)
{\em Learning to forget: continual prediction with LSTM},
Neural Computation, 12(10):2451--2471

\bibitem{GRU} 
Kyunghyun Cho et al. (2014)
{\em Learning Phrase Representations using RNN Encoder–Decoder for Statistical Machine Translation},
Proceedings of the 2014 Conference on Empirical Methods in Natural Language Processing (EMNLP), pages 1724--1734

\bibitem{Elman}
Elman, J. L. (1990). {\em Finding structure in time}. 
Cognitive Science, 14(2), 179–211.

\bibitem{transformer}
Vaswani, A., Shazeer, N., Parmar, N., Uszkoreit, J., Jones, L., Gomez, A. N., Kaiser, L., and Polosukhin, I. (2017). Attention Is All You Need. Advances in Neural Information Processing Systems, 30, 5998–6008.

\bibitem{Shaw}
Peter Shaw, Jakob Uszkoreit, and Ashish Vaswani (2018)
{\em Self-Attention with Relative Position Representations},
Proceedings of the 2018 Conference of the North American Chapter of the Association for Computational Linguistics (NAACL), 464--468


\bibitem{Evaluation of NLPs}
G Melis, C Dyer, and  P Blunsom (2017)
{\em On the state of the art of evaluation in neural language models}, 
arXiv preprint arXiv:1707.05589

\bibitem{Wu2016}
Yuhuai Wu, et al. (2016)
{\em On multiplicative integration with recurrent neural networks}, 
CoRR, abs/1606.06630, 2016. http://arxiv.org/abs/1606.06630.

\bibitem{Krauser}
Krause, B., Lu, L., Murray, I., and Renals, S. (2016) 
{\em Multiplicative LSTM for sequence modelling. }
arXiv:1609.07959

\bibitem{cooijmans}
Tim Cooijmans et al. (2016)
{\em Recurrent Batch Normalization},
arXiv:1603.09025

\bibitem{Chung2016}
Junyoung Chung, Sungjin Ahn, Yoshua Bengio (2016)
{\em Hierarchical Multiscale Recurrent Neural Networks},
arXiv:1609.01704
	
\bibitem{Chung}
Junyoung Chung, Sungjin Ahn, and Yoshua Bengio (2016) 
{\em Hierarchical multiscale recurrent neural networks},
 CoRR, abs/1609.01704, http://arxiv.org/abs/1609.01704.


\bibitem{HLSH}
Moses S. Charikar (2002) Similarity Estimation Techniques from Rounding
Algorithms, {\em Proceedings of the thiry-fourth annual ACM symposium on Theory of computing}, 380--388

\bibitem{Toy Model of Superposition}
Nelson Elhage et al. (2022) 
{\em Toy Models of Superposition}, 
https://transformer-circuits.pub/2022/toy\_model

\bibitem{toy-double-descent}
Tom Henighan et al. (2023)
{\em Superposition, Memorization, and Double Descent}
https://transformer-circuits.pub/2023/toy-double-descent

\bibitem{Neurobiology}
Shepherd, G. M. (1988). Neurobiology (2nd ed.). Oxford University Press. 

\bibitem{Rate vs Temporal}
Jacques Gautrais and Simon Thorpe (1998)
{\em Rate coding versus temporal order coding: a theoretical approach}. 
Biosystems, 48(1-3):57-65.

\bibitem{SimonThorpe}
Simon Thorpe, Arnaud Delorme, and Rufin Van Rullen (2001)
{\em Spike-based strategies for rapid processing},
Neural Networks, 14: 715-725

\bibitem{backpropinthebrain}
Lillicrap, T. P., Santoro, A., Marris, L., Akerman, C. J., and  Hinton, G. (2020). Backpropagation and the brain. Nature Reviews Neuroscience, 21(6), 335–346

\bibitem{Reformer}
Nikita Kitaev, Lukasz Kaiser, and Anselm Levskaya (2020)
Reformer: The Efficient Transformer, 
arXiv:2001.04451

\bibitem{LoRA}
Edward J. Hu et al. (2021) \emph{LoRA: Low-Rank Adaptation of Large Language Models}, ArXiv, 2106.09685


\bibitem{Shazeer}
Noam Shazeer at al. (2017)
{\em Outrageously large neural networks: The sparsely-gated mixture-of-experts layer}, 
arXiv:1701.06538

\bibitem{Sparse MoE}
William Fedus, Jeff Dean, and Barret Zoph (2022)
{\em A Review of Sparse Expert Models in Deep Learning},
arXiv:2209.01667


\bibitem{RandomForest}
Breiman, L. (2001). Random Forests. Machine Learning, 45(1), 5-32.

\bibitem{RandomFern}
M. Ozuysal, M. Calonder, V. Lepetit and P. Fua, (2010) {\em Fast Keypoint Recognition Using Random Ferns,} in IEEE Transactions on Pattern Analysis and Machine Intelligence, vol. 32, no. 3, pp. 448-461

\bibitem{DeepForest}
Z. H. Zhou and J. Feng. "Deep Forest." National Science Review, 2019, 6(1): 74-86

\bibitem{DeepFerns}
Kim, S., and Ko, B. C. (2020). Building Deep Random Ferns Without Backpropagation. IEEE Access, 8, 8956-8965.


\bibitem{SpikeDriven}
Man Yao et al. (2023)
Spike-driven Transformer,
arXiv:2307.01694v1

\bibitem{SpikeLLM}
Xingrun Xing, Boyan Gao, Zheng Zhang, David A. Clifton, Shitao Xiao, Li Du, Guoqi Li, Jiajun Zhang (2024) 
SpikeLLM: Scaling up Spiking Neural Network to Large Language Models via Saliency-based Spiking.
arXiv:2407.04752


\bibitem{SpikeGPT}
Rui-Jie Zhu, Qihang Zhao, Guoqi Li, Jason K. Eshraghian (2024)
SpikeGPT: Generative Pre-trained Language Model with Spiking Neural Networks. arXiv:2302.13939

\bibitem{SpikeBert}
Changze Lv, Tianlong Li, Jianhan Xu, Chenxi Gu, Zixuan Ling, Cenyuan Zhang, Xiaoqing Zheng, Xuanjing Huang (2024)
SpikeBERT: A Language Spikformer Learned from BERT with Knowledge Distillation. 
arXiv:2308.15122

\bibitem{SpikeFormer}
Yudong Li, Yunlin Lei, and Xu Yang. (2024)
Spikeformer: A Novel Architecture for Training High-Performance Low-Latency Spiking Neural Network.
Neurocomputing, Volume 574, March 14

\bibitem{BrainTransformers}
Zhengzheng Tang, Eva Zhu (2024)
BrainTransformers: SNN-LLM. 
arXiv:2410.14687 


\bibitem{SpikeBrain}
Yuqi Pan et al (2025)
SpikingBrain Technical Report: Spiking Brain-inspired Large Models,
arXiv:2509.05276

\bibitem{LLMonLoihi2}
Steven Abreu, Sumit Bam Shrestha, Rui-Jie Zhu, Jason Eshraghian (2025)
{\em Neuromorphic Principles for Efficient Large Language Models on Intel Loihi 2},
arXiv:2503.18002


\bibitem{MatMulFree}
Rui-Jie Zhu, Yu Zhang, Steven Abreu, Ethan Sifferman, Tyler Sheaves, Yiqiao Wang, Dustin Richmond, Sumit Bam Shrestha, Peng Zhou, Jason K. Eshraghian
{\em Scalable MatMul-free Language Modeling},
arXiv:2406.02528


\bibitem{AlexNet}
Alex Krizhevsky, Ilya Sutskever, Geoffrey E. Hinton (2012) 
{\em ImageNet Classification with Deep Convolutional Neural Networks}, 
 Advances in Neural Information Processing Systems 25 (NIPS 2012)


\end{thebibliography}
\end{document}